\documentclass[journal]{IEEEtran}
\IEEEoverridecommandlockouts                     

\usepackage[free-standing-units=true,binary-units=true]{siunitx}
\usepackage{graphicx}
\usepackage[ansinew]{inputenc}
\usepackage{amssymb, amsmath, amsthm}
\usepackage{stmaryrd}
\usepackage{subcaption}
\usepackage{paralist, tabularx}
\usepackage{xspace}
\usepackage{enumitem}
\usepackage{multirow}
\usepackage{xcolor,colortbl}
\usepackage{mathtools}
\usepackage[pstricks1-10]{vaucanson-g}
\newcommand\Loadedframemethod{default}
\usepackage[framemethod=\Loadedframemethod]{mdframed}

\usepackage[font=small]{caption}
\newboolean{showcomments}
\setboolean{showcomments}{true} 

\ifthenelse{\boolean{showcomments}}
{
	\newcommand{\del}[1]{\textcolor{red}{\sout{#1}}}    
}{
	\newcommand{\del}[1]{}                              
	
}

\ifthenelse{\boolean{showcomments}}{
	\newcommand{\nbc}[3]{
		{\colorbox{#3}{\bfseries\sffamily\scriptsize\textcolor{white}{#1}}}
		{\textcolor{#3}{\sf\small$\langle$\textit{#2}$\rangle$}}}
}{
	\newcommand{\nbc}[3]{}
	
}

\newcommand{\formatAction}[1]{{\footnotesize \textsf{#1}}\xspace}
\newcommand{\formatFluent}[1]{{\small \emph{#1}}\xspace}

\newcommand{\lig}[2]{\lightning_{\!\!#1}^{\!#2}}

\mdfsetup{skipabove=\topskip,skipbelow=\topskip}
\mdfdefinestyle{myframe}{%
    linecolor=black, %
    outerlinewidth=2pt, %
    roundcorner=20pt,
    innertopmargin=4pt, %
    innerbottommargin=4pt, %
    innerrightmargin=4pt, %
    innerleftmargin=4pt, %
        leftmargin = 4pt, %
        rightmargin = 4pt, %
        }

\newcommand{\paragraphTheme}[1]{} 

\newcommand{\plan}[1]{{\small \textit{\textsf{Plan\_#1}}}}

\newcommand{\lowHeight}{\formatAction{low.height}}
\newcommand{\highHeight}{\formatAction{high.height}}

\newcommand{\hotSwap}{\formatAction{hotSwap}}
\newcommand{\stopOld}{\formatAction{stopOld}}
\newcommand{\startNew}{\formatAction{startNew}}
\newcommand{\reconfigure}{\formatAction{reconfig}}
\newcommand{\stopOldfl}{\formatFluent{OldStopped}}
\newcommand{\startNewfl}{\formatFluent{NewStarted}}
\newcommand{\HotSwap}{\formatFluent{HotSwap}}

\newcommand{\Reconfigurefl}{\formatFluent{Reconfigured}}

\newcommand{\senseormove}{\formatFluent{SenseOrMoveCmd}}
\newcommand{\iterempty}{\formatFluent{Reset}}

\newcommand{\hasNext}{\formatAction{has.next?}}
\newcommand{\remove}{\formatAction{remove.next}}
\newcommand{\reset}{\formatAction{reset}}
\newcommand{\yesNext}{\formatAction{y.next}}
\newcommand{\noNext}{\formatAction{n.next}}
\newcommand{\takeOff}{\formatAction{takeOff}}

\renewcommand{\land}{\formatAction{land}}

\newcommand{\go}{\formatAction{go}}
\newcommand{\goN}[1]{\formatAction{go.#1}}

\newcommand{\spin}{\formatAction{do.spin}}
\newcommand{\spinend}{\formatAction{spin.ended}}
\newcommand{\grab}[1]{\formatAction{grab.#1}}
\newcommand{\release}[1]{\formatAction{release.#1}}

\newcommand{\inP}{\formatAction{is.next.inA?}}
\newcommand{\yinP}{\formatAction{yes.next.inA}}
\newcommand{\ninP}{\formatAction{no.next.inA}}
\newcommand{\sensor}[1]{\emph{sensor.\region{#1}}}

\newcommand{\region}[1]{\textit{\textsc{#1}}}
\newcommand{\old}{{\small\textsc{OldSpec}}\xspace}
\newcommand{\new}{{\small\textsc{NewSpec}}\xspace}
\newcommand{\with}[1]{\formatFluent{Carrying.#1}}

\newcommand{\Gl}{\Box}
\newcommand{\F}{\Diamond}

\newcommand{\W}{\; \mbox{W} \;}

\newcommand{\set}[1]{\{ #1 \}}

\newcommand{\gr}{{GR(1)}\xspace}

\newcommand{\arrivedfl}[1]{\formatFluent{At.#1}}

\newcommand{\arrivedN}[1]{\formatAction{at.#1}}
\newcommand{\photofl}{\formatFluent{Img.processed}}

\newcommand{\nfold}{NoFlyOld\xspace}
\newcommand{\nfnew}{NoFlyNew\xspace}

\newcommand{\Journal}[1]{}

\newcommand{\modelsF}[1]{\ensuremath{\models}}

\newcommand{\morph}[1]{\textit{#1}\xspace}
\newcommand{\swarch}[1]{\textit{#1}\xspace}

\newcommand{\morphSynthesiser}{\morph{Synthesiser}}
\newcommand{\morphCtrlManagment}{\morph{Controller Management}}

\newcommand{\morphHybridControl}{\morph{Hybrid Control}}

\newcommand{\morphVehicle}{\morph{Vehicle}}
\newcommand{\morphOfftheshelf}{\morph{Off-the-shelf Vehicle}}
\newcommand{\morphComputer}{\morph{Onboard Computer}}
\newcommand{\morphGCS}{\morph{Ground Control Station}}

\begin{document}


\title{Assured Mission Adaptation of UAVs}


\author{Sebasti\'an A. Zudaire$^{1}$,
Leandro Nahabedian$^{2}$,
and Sebasti\'an Uchitel$^{3}$%
\thanks{$^{1}$Instituto Balseiro - Universidad Nacional de Cuyo, Argentina.
        {\tt\small sebastian.zudaire@ib.edu.ar}}%
 \thanks{$^{2}$Universidad de Buenos Aires / CONICET, Argentina
                {\tt\small lnahabedian@dc.uba.ar}}%
\thanks{$^{3}$Universidad de Buenos Aires,
Buenos Aires, Argentina and Imperial College London, UK.
        {\tt\small suchitel@dc.uba.ar}}%
}

\maketitle
\begin{abstract}

The design of systems that can change their behaviour to account for scenarios that were 
not foreseen at design time remains an open challenge. 
In this paper we propose an approach for adaptation of mobile robot 
missions that is not constrained to a predefined set of mission evolutions. We 
propose 
applying the MORPH adaptive software architecture to UAVs and 
show how controller synthesis can be used both to guarantee correct transitioning 
from the old to the new mission goals while architectural reconfiguration to include 
new software actuators 
and sensors if necessary. The architecture brings together architectural concepts that 
are commonplace in robotics such as temporal planning, discrete, hybrid and 
continuous control layers 
together with architectural concepts from adaptive systems such as runtime models 
and runtime synthesis. We validate the architecture flying several missions taken from the 
robotic literature for different real and simulated UAVs.

\end{abstract}




\section{Introduction} \label{sec:intro}

Adaptive systems are capable of changing their 
behaviour while running in response to changes in their 
environment, capabilities and goals~\cite{de2013software}. 
Adaptation can be addressed at various levels of 
abstraction to respond to many different 
kinds of changes. In this paper we focus on mobile robot adaptations that involve 
responding to unforeseen changes in the high-level goals that the robot 
must achieve.

%
%

Consider an Unmanned Aerial Vehicle (UAV) that is performing a remote patrol 
mission, 
flying at a high 
altitude between series of patrol points, recording the 
ground with a camera to 
relay a low-res movie through a low-bandwidth 
channel back to a control centre. While in 
mission, a report comes into the control centre that a 
person, known to be 
wearing a red jacket, has gone missing near the patrol 
area. 

Rather than flying the UAV to base, programming a 
search mission for the UAV, deploying new software 
and sending the UAV back to the original patrol area, it 
would be convenient to have the UAV designed to 
support the following scenario. 

While in flight, personnel at the control centre specify a 
new mission for the UAV that involves systematically 
searching the area at low altitude using 
a red-sensor detection filter on high-res photos, 
landing on detection. The team 
prepares for upload an image 
processing module. They also specify key mission 
transition requirements: The module 
cannot be bound into the software while 
the camera is in use and the camera must be realigned 
to point downwards before the new mission is started. 

Having produced the specifications and new software 
modules, mission command pushes a button and code 
is automatically synthesised to satisfy the transition 
requirements and the new mission. The synthesised 
code is uploaded onto the UAV and starts running. The 
current mission is stopped, the camera realigned and 
set to high-res mode, the new flight altitude is 
configured and then, ensuring that the camera is not in 
use, the uploaded image processing module is bound 
into the architecture. Once the UAV reaches the new 
altitude, the search mission commences.  

In this paper we explore the question of how UAV systems can be designed 
to support adapting to unforseen circumstances by changing missions at fly-time.
Our hypothesis is that discrete event controller synthesis at runtime can help provide 
a flexible mission adaptation mechanism with guarantees not only about satisfying the 
new mission requirements but also safely transitioning between the current and the 
new mission requirements. 


We report on a robotic system that supports assured runtime adaptation of missions.
The mission (specified as a combination of automata and temporal logic 
formulae) can involve re-discretization of the robot workspace and reconfiguration of 
the robot's software architecture, with changes in the available software sensors and 
actuators. Correctness criteria for the mission transition and reconfiguration is  
provided by the user as a temporal logic formula. 
The system relies on discrete event-controller synthesis to produce a plan that safely 
reconfigures and transitions into the new mission. The plan is executed on a hybrid 
control architecture that supports runtime swapping of plans, and runtime binding and 
unbinding of hybrid components. 

\subsection{Related Work} \label{sec:relwork}
\textit{Runtime change of software systems} has been studied extensively. Different 
application domains and technology stack pose different problems and require 
different solutions~\cite{SMR:SMR1556}.
A major concern is  \textit{correctness preservation} throughout  change. Many 
approaches assume that there is no change in the intended system behaviour (i.e., 
the \textit{specification/mission remains unchanged}) and that a patch is being 
applied (e.g.,~\cite{Hosek:2013:SSU:2486788.2486869}). Alternatively, a set of 
\textit{fixed domain independent properties}, such as consistency, are expected to hold 
(e.g.,~\cite{DSURelaxedConsistencyModel,journals/tse/GuptaJB96,
Kramer:1990:EPP:93658.93672,Banno10}). More recently, complex plans for 
supporting architectural change while preserving user provided structural constraints 
has been studied (e.g.,~\cite{Tajalli2010}).
Some approaches do support domain specific specification changes; however, they 
require a \textit{prespecified universe of possible changes} at the time of running the 
system 
for the first time (e.g.,~\cite{schmidt}).

The need for supporting arbitrary specification changes and \textit{update 
requirements} that constrain the transition between specifications 
has been subject of more recent studies 
(e.g.,\cite{Hayden:2012:SVC:2189314.2189336,Ramirez:2010, 
Zhang:ICSE2006}). 
Although they focus on \textit{specification and 
verification of update strategies}, more recently work focuses on \textit{automated 
synthesis of update 
plans} 
(e.g.,~\cite{Baresi:2010:DBD:1882362.1882367,LaManna2013,Nahabedian-TSE}).

Existing work in this area, however, is \textit{insufficiently expressive}  to 
accommodate the liveness requirements that typical robotic missions 
have~\cite{Menghi19}. Furthermore, existing approaches do not address the specifics 
of temporal mission planning of mobile robots~\cite{Belta07,Fainekos05} 
including 
changes to sensor 
and actuator abstractions, including changes in discretization. Such changes require 
reasoning not only about when and how to change the system behaviour but also 
when to introduce \textit{software reconfiguration}  (e.g., binding and unbinding new 
software components). A proposal for the latter has been outlined in the 
MORPH reference 
architecture~\cite{MORPH15} but does not define mechanisms for ensuring correct 
adaptation nor does it resolve the non-trivial specifics of 
applying these ideas to a hybrid control architecture~\cite{Fainekos05,Gazit08} required 
to resolve 
the discrete-continuous gap between 
mission specifications and the physical world.

%
%
%


\subsection{Summary}
The main contributions of this paper  is a  system  for adapting UAV 
missions with correctness guarantees that a) builds on discrete event controller 
update~\cite{Nahabedian-TSE} but extends it to liveness properties to support 
typical mobile robot 
missions~\cite{Menghi19}, b) that implements a hybrid controller~\cite{Gazit08}
architecture that incorporates reconfiguration capabilities 
from~\cite{MORPH15}. We demonstrate, in real and simulated flights, how a UAV 
running a mission can be adapted at runtime to new missions that may require 
changes to 
workspace  discretization, software sensors and software actuators.




In Section~\ref{sec:dcu4robots} we show how UAV missions and mission 
adaptations can be specified. This sets the appropriate level of abstraction to present 
an overview of the architecture (Section~\ref{sec:extension}) and 
detailed description of the software architecture and main software components when 
applied to UAVs (Section~\ref{subsec:implementation}). We then report on our validation 
efforts in Section~\ref{sec:Validation} and conclude with a discussion and future work 
(Section~\ref{sec:discussion}).


\section{Preliminaries}
\label{sec:Preliminaries}


\subsection{Labelled Transition Systems (LTS)}

The dynamics of the interaction of the robot with its environment are
modelled using LTS~\cite{Keller76}, which are automata where
transitions are labelled with events that constitute the interactions
of the modelled system with its environment.
We partition events into controlled and uncontrolled to specify
assumptions about the environment and safety requirements for a
controller.
Complex models can be constructed by LTS composition. We use a
standard definition of \textit{parallel composition} ($\|$) that models the
asynchronous execution of LTS, interleaving non-shared actions and
forcing synchronisation of shared actions. We use an \textit{interrupt 
operator}~\cite{Nahabedian-TSE} ($E 
\lig{f}{\ell} E'$) to model that the behaviour described by LTS $E$ may be 
interrupted 
by event $\ell$ to become LTS $E'$. Function $f$ sets the initial state of $E'$ based on 
the state of $E$ when the interrupt happens.

\subsection{Fluent Linear Temporal Logic (FLTL)}
In order to describe environment assumptions and system goals it is
common to use formal languages like FLTL~\cite{gianna03}, a variant
of linear-time temporal logic that uses fluents to describe states
over sequences of actions.
A fluent $\formatFluent{fl} = \langle Set_\top, Set_\bot, v \rangle$ is defined by a set 
of initiating actions ($Set_\top$),
a set of terminating actions ($Set_\bot$), and an initial value $v$ true ($\top$) or 
false ($\bot$).
We may omit set notation for singletons and use an action label $\ell$
for the fluent defined as
$\formatFluent{fl} = \langle \ell, \emph{Act}\setminus\set{\ell}, \bot
\rangle$. Thus, the fluent $\ell$ is only true just after the
occurrence of the action $\ell$.
FLTL is defined similarly to propositional LTL but where a fluent
holds at a position $i$ in a trace $\pi$ based on the events occurring
in $\pi$ up to $i$. Temporal connectives are interpreted as usual:
$\F \varphi$, $\Gl \varphi$, and $ \varphi \W \psi$ mean that $\varphi$
eventually holds, always holds, and (weakly) holds until $\psi$, respectively. An LTS 
$E$ satisfies $\varphi$  ($E \models \varphi$) when all its traces satisfy $\varphi$. 
We refer to liveness formulae as those that only have infinite trace violations. 
Otherwise we refer to them as safety formulae. 

\subsection{Discrete Event Controller Synthesis}
\label{pre:CS}
We adopt the controller synthesis formulation from~\cite{dippolito11}. Given an LTS 
$E$ describing the execution environment of a discrete controller with a 
set of controllable actions $L$ and a task specification 
$\varphi$ 
expressed in FLTL, the goal of controller synthesis is
to find an LTS $C$ such that $E\|C$: (1) is deadlock free, (2) $C$
does not block any non-controlled actions, and (3) 
$E\|C \models \varphi$ 
 We say that a control problem $\langle E, \varphi, L \rangle$ is 
\textit{realizable} if 
such an LTS 
$C$ exists.
The tractability of the controller synthesis depends on the size of the problem (i.e. 
states of $E$ and size of $\varphi$) and also on the fragment of the logic used for 
$\varphi$.
When goals are restricted to safety formulae and \gr formulae the control problem 
can be solved in polynomial
time~\cite{piterman06}. \gr formulas are of the form $\bigwedge_{i=1}^n \Gl\F A_i 
\implies 
\bigwedge_{i=1}^m \Gl\F G_i$  where $Ai_i$ and $G_i$
  are Boolean combinations of fluents.
In this paper we use MTSA~\cite{MTSA} for solving control problems.

\subsection{Dynamic Controller Update} \label{sec:prelimDCU}
We summarise the results in~\cite{Nahabedian-TSE}.
Assume a controller $C$ that is solution to a control problem $\langle E, \varphi, L 
\rangle$ that is to be replaced by a new controller $C'$, with $\varphi = \Gl G$ and 
$G$ a boolean 
combination of fluents. The term $(C\lig{f}{\hotSwap} 
C')$ models hot-swapping one controller with the other, where $C'$ is initialised based 
on the current state of $C$ at the time of \hotSwap using function $f$. Event 
\hotSwap is uncontrolled. 

Assume the intention is that the new controller $C'$  guarantees $\varphi'$ in a
new execution environment $E'$, with $\varphi' = \Gl G'$ and $G'$ a boolean 
combination of fluents. In 
addition, consider a function $g$ that initialises 
the state of $E'$ based on the current state of $E$ and a controlled event 
$\reconfigure$ that models the reconfiguration of the 
execution environment $E$ by $E'$, i.e., $E\lig{g}{\reconfigure} E'$.

In addition, we introduce two more controlled events: $\stopOld$ that signals from 
when $\varphi$ is no longer 
guaranteed (i.e., $G \; \mathbf{W} \; \stopOld$), and $\startNew$ that signals 
from when $\varphi'$ is 
guaranteed (i.e., $\Gl (\startNew \implies 
\varphi')$).

Finally, assume a safety FLTL formula $\Theta$ that models the transition requirement 
between 
controllers. In other words, constrains the occurrence of $\reconfigure$, $\startNew$ 
and $\stopOld$.  As an example consider a standard domain independent 
requirement $\Theta_\emptyset = \Gl (\neg \stopOldfl \vee \startNewfl)$, with 
fluents 
\stopOldfl and \startNewfl turning on 
with the occurrence of \stopOld and \startNew, respectively, and never turning off. 
$\Theta_\emptyset$ states that the system must always be under control to achieve 
the 
old specification ($\varphi$) or the new one ($\varphi'$).

%

The dynamic controller update problem is to find $C'$ and $f$ such that (1) $f$ is a 
total function, (2)  $C\lig{f}{\hotSwap} C'$ does not block 
any non-controlled actions in $E\lig{g}{\reconfigure} E'$, and that $C\lig{f}{\hotSwap} 
C' 
\|  E\lig{g}{\reconfigure} E'$ is (3) deadlock free, and (4) satisfies $\varphi_{DCU}$ 
defined as the conjunction of the following:
\begin{compactenum}
\item $G \; \mathbf{W} \; \stopOld$
\item $\Theta$
\item $\Gl (\startNew \implies \varphi')$
\item  $\Gl (\hotSwap \implies (\F \stopOld \; \wedge \F 
\reconfigure \; 
\wedge \; \F 
\startNew)) $
\end{compactenum}

Note that conjunct (4) of $\varphi_{DCU}$ requires that once the uncontrolled event 
\hotSwap occurs, the system must eventually reconfigure and switch specifications.

In~\cite{Nahabedian-TSE}, DCU problem is reduced to a 
standard discrete event control problem, and tool (MTSA~\cite{MTSA}) is reported 
that computes an LTS $C_u$ of the form for  $C\lig{f}{\hotSwap} C'$ from which $C'$ 
and $f$ can be extracted.

\subsection{Dynamic Controller Update of Live Missions} \label{sec:prelimDCULive}

The DCU problem in~\cite{Nahabedian-TSE} is restricted to safety requirements.  
However, a typical mobile 
robot mission will have liveness properties. For instance, and taking patterns 
recollected from scientific literature and industrial case studies~\cite{Menghi19},
patrolling two locations requires $\Gl \F \arrivedN{1} \wedge \Gl \F \arrivedN{2}$ 
and a delivery mission requires $\Gl (\formatAction{available} \implies \F 
\formatAction{deliver})$.

Simply lifting the safety restriction over $\varphi$ and $\varphi'$ leads to problems. 
Consider $\varphi = \Gl (\formatAction{available} \implies \F 
\formatAction{deliver})$ and the requirement that the update must satisfy:
$(\formatAction{available} \implies \F 
\formatAction{deliver}) \W \stopOld$.  Such a requirement effectively does not allow 
scenarios in which it is necessary to abort a pending delivery 
 obligations. 

To provide a more general framework for updating controllers, we redefine the 
property 
$\varphi_{DCU}$ that $C\lig{f}{\hotSwap} 
C' 
\|  E\lig{g}{\reconfigure} E'$ is expected to satisfy.  We assume without loss of 
generality that a 
mission specifications ($\varphi$) is split into a safety part ($\varphi_S)$ and  a 
liveness part $\varphi_L$. We define  $\varphi_{DCU_L}$ as:

\begin{compactenum}
\item $\varphi_S \; \mathbf{W} \; \stopOld$
\item $\Theta$
\item $\Gl \big(\startNew \implies \varphi' \big)$
\item $\Gl (\hotSwap \implies (\F \stopOld \; \wedge \F \reconfigure \; \wedge \; 
\F  \startNew))$
\end{compactenum}

Note that in (1)  there is no mention of the liveness part of the old specification 
($\varphi_L$), consequently all liveness obligations will be dropped as soon as the 
new controller is put in place. Should this not be desired, $\Theta$ 
can be defined to 
prevent this (e.g., $ \Theta = ((\formatAction{available} \implies \F 
\formatAction{deliver}) \W \stopOld)$. 

If $\varphi' = \varphi_S \wedge \varphi_L$ and $\varphi_L$ is of the form 
$\bigwedge_{i=1}^m \Gl\F G_i$ then items (3) and (4) can be combined to conform 
to an equivalent safety plus GR(1) formula: 
$\varphi_S' \wedge [\Gl\F \HotSwap \implies \Gl \F(\stopOldfl \wedge \startNewfl 
\wedge 
\Reconfigurefl)
\wedge \varphi_L']$, where $\stopOldfl$, $\startNewfl$, and 
$\Reconfigurefl$ are fluents initially false that become true once $\stopOld$, 
$\startNew$ and $\reconfigure$ occur. 
Thus,  
a similar pattern as described 
in~\cite{Nahabedian-TSE} can be applied to polynomially solve DCU problems with 
recurrent liveness 
missions (i.e., $\bigwedge_{i=1}^m \Gl\F G_i$). 
We have extended the implementation of the MTSA~\cite{MTSA} tool to support 
solving these control problems. 

\section{Specification and Synthesis of Assured Adaptation Plans} 
\label{sec:dcu4robots}

In this section we show how assured mission adaptation of mobile robots can be 
framed as a Dynamic Controller Update (DCU) problem. A solution to the DCU problem 
yields a controller that codifies a plan that ensures the mobile robot will change its 
current mission plan to a plan that satisfies its new mission. 
In the next section we discuss a hybrid control architecture that can make use of a 
solution 
to the DCU problem to actually adapt a UAV at runtime. 

We first show how a simple case mobile robot mission adaptation can 
solved using DCU.  We illustrate the importance of obtaining one controller that 
works for any reachable state of the current mission plan (i.e.  $C\lig{f}{\hotSwap} 
C'$).
Nonetheless, this example is simple in that the term $E\lig{g}{\reconfigure} E'$ of the 
DCU control problem allows $E = E'$ and $g$ be the identity function. The transition 
requirement $\Theta$ also plays a minor role. 

The second example discusses a mission adaptation that requires introducing new 
software components and re-discretization of the robot's workspace. For this, the 
DCU problem requires $E \neq E'$ and an appropriate mapping function $g$.  
The third example, shows the relevance of $\Theta$ to solve adaptation problems in 
which the new and old missions are logically inconsistent. 
%


\subsection{Adaptation of Live Missions} \label{subsec:adaplive}

\begin{figure*}
\begin{subfigure} {0.32\linewidth}
\centering
\includegraphics[width=\linewidth]{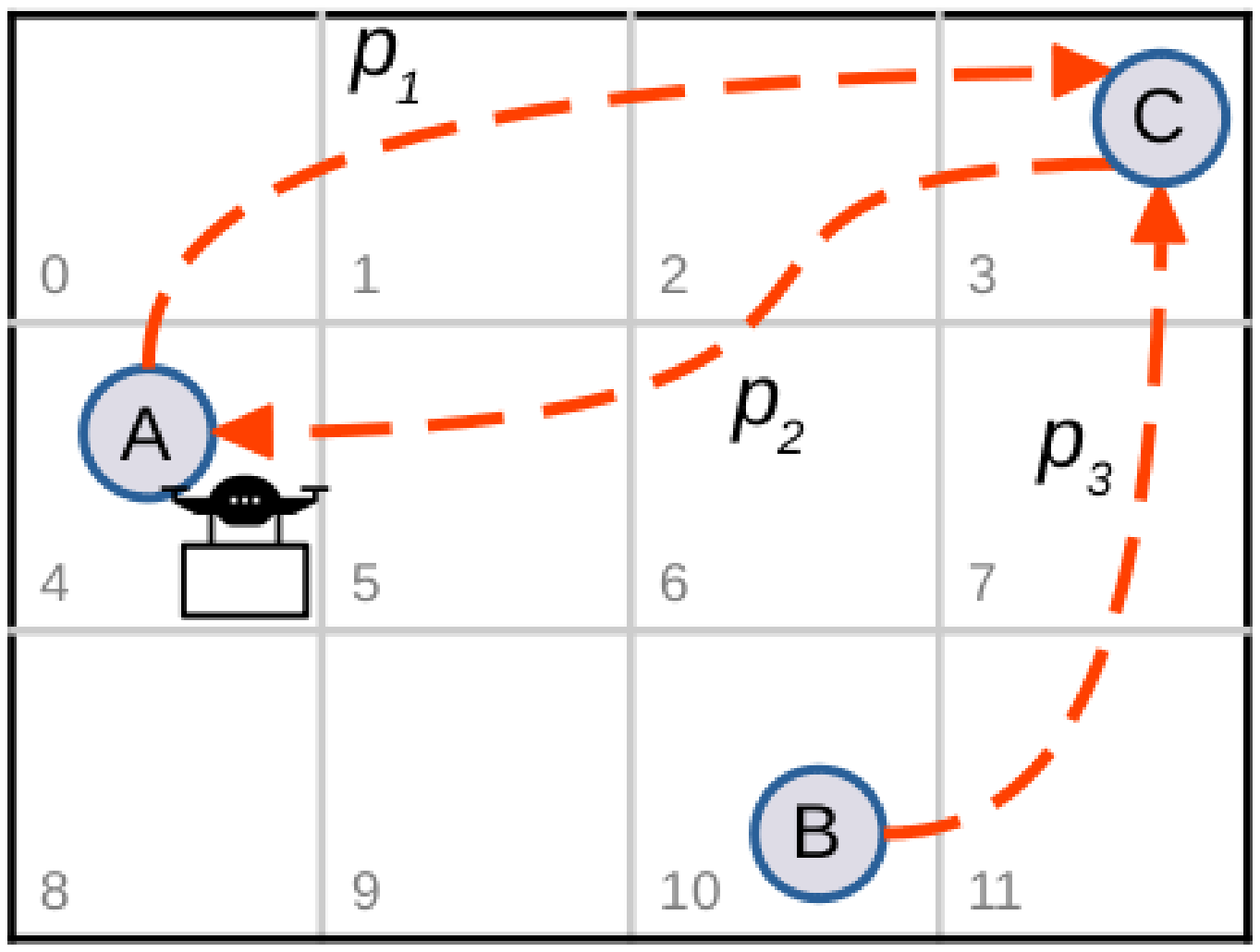}
\caption{Original delivery requirements and initial UAV location.}
\label{fig:delivery1}
\end{subfigure}
~
\begin{subfigure} {0.32\linewidth}
\centering
\includegraphics[width=\linewidth]{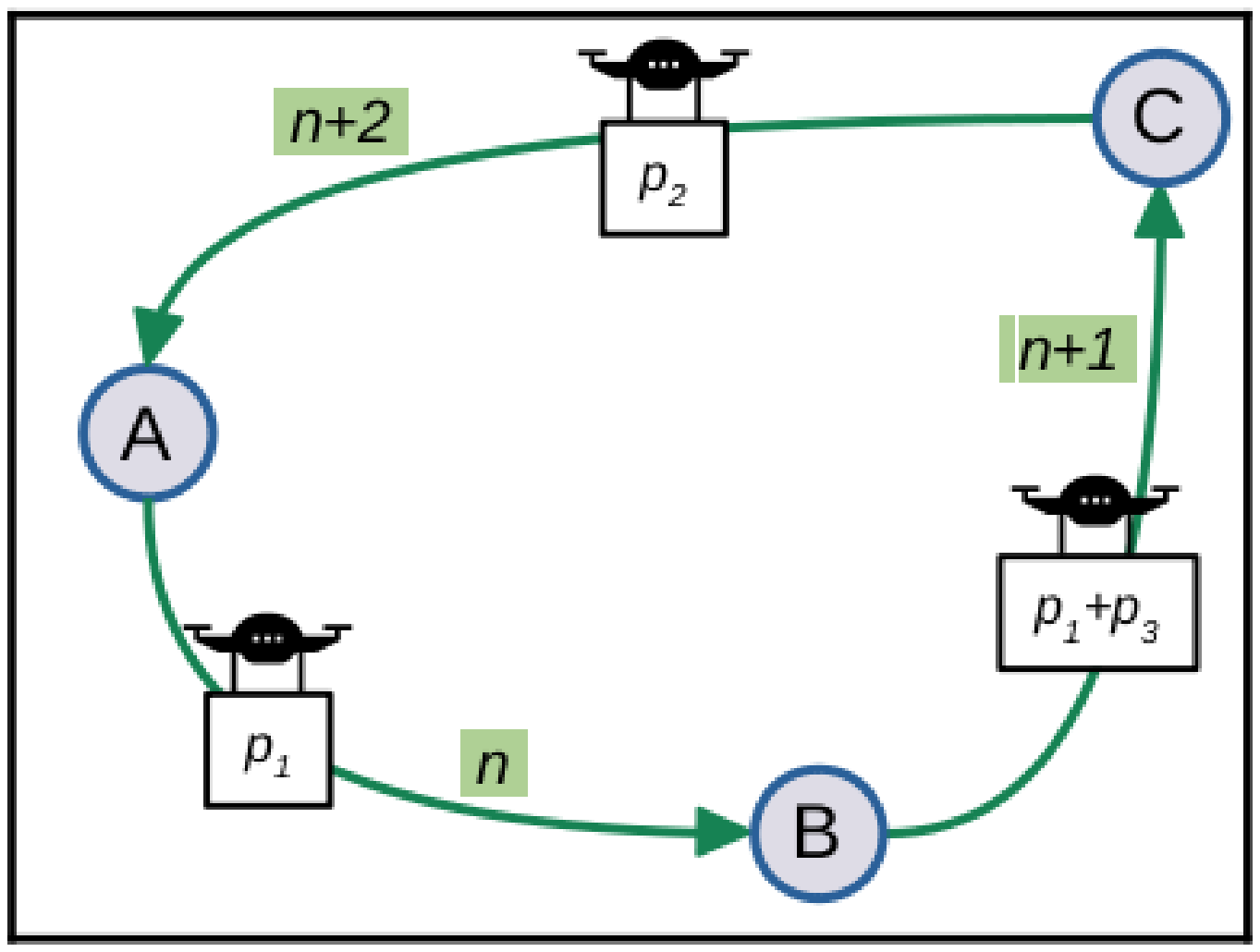}
\caption{Original delivery plan, showing the UAV correctly delivering each of the 
packets.}
\label{fig:delivery1sol}
\end{subfigure}
~
\begin{subfigure} {0.32\linewidth}
\centering
\includegraphics[width=\linewidth]{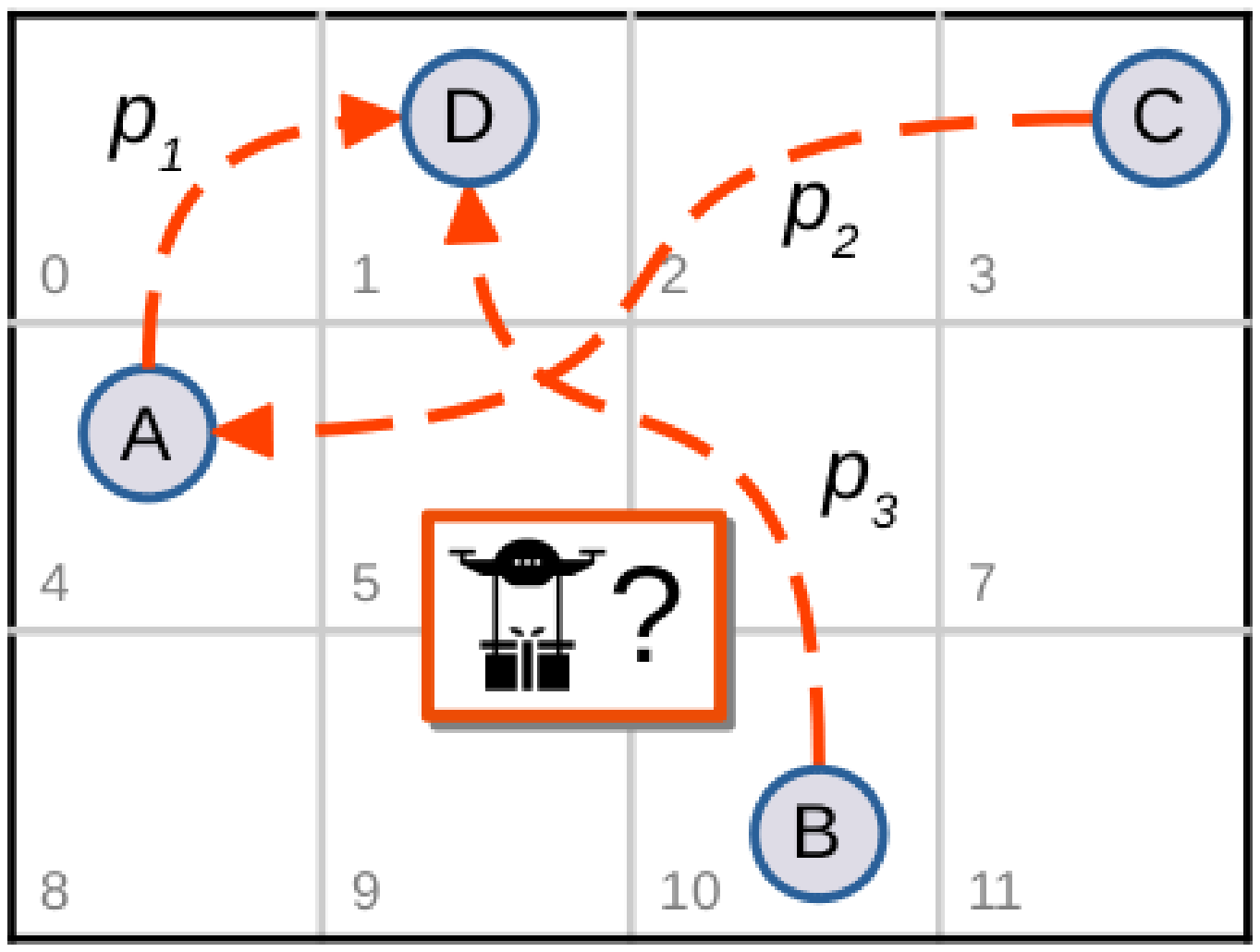}
\caption{New delivery requirements with new drop off location \region{D}, initial UAV 
position unknown.}
\label{fig:delivery2}
\end{subfigure}
~
\begin{subfigure} {0.32\linewidth}
\centering
\includegraphics[width=\linewidth]{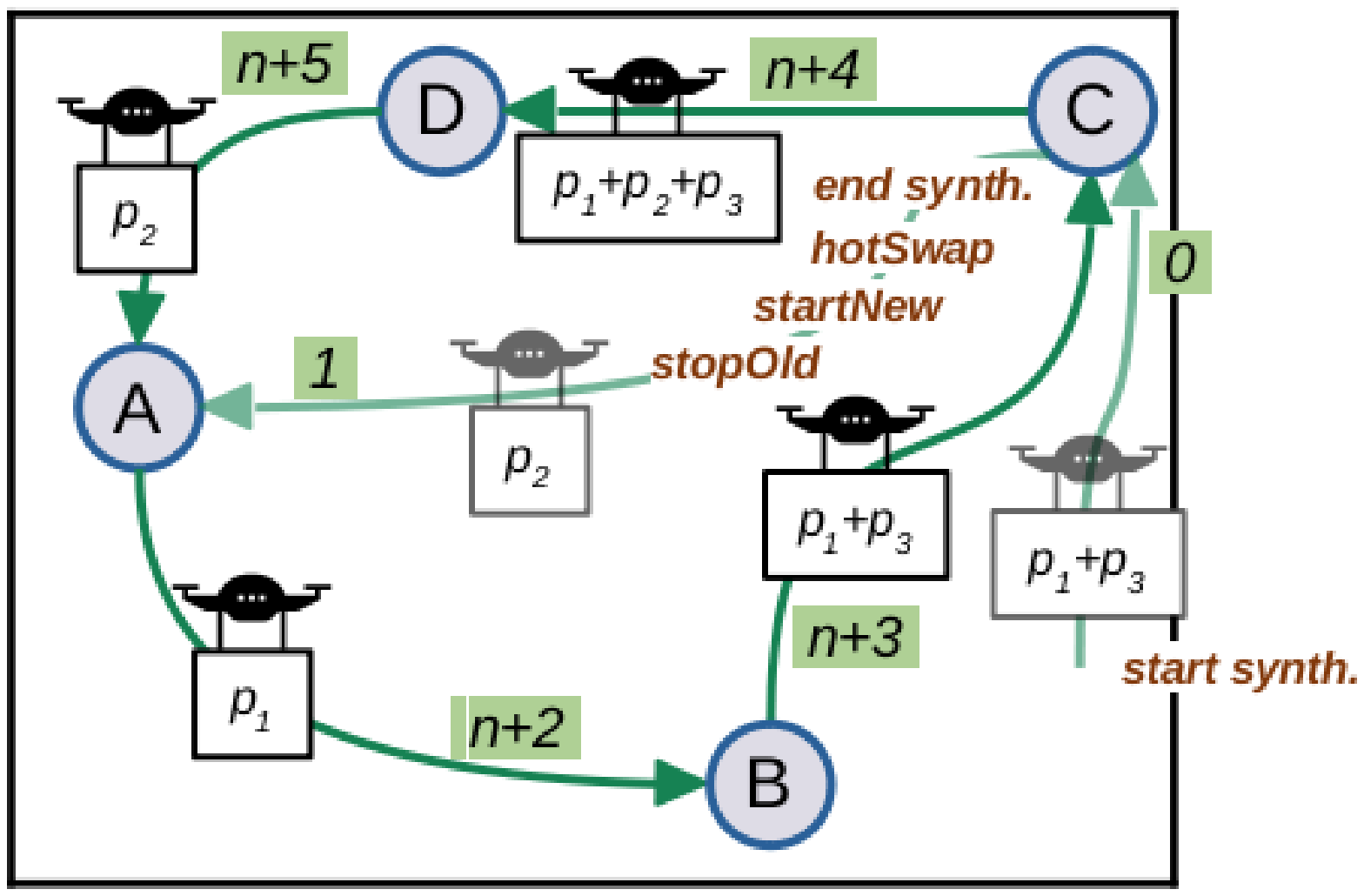}
\caption{Update plan when the \hotSwap occurs between \region{C} and \region{A}, while 
carrying $p_2$.}
\label{fig:delivery2sol2}
\end{subfigure}
~
\begin{subfigure} {0.32\linewidth}
\centering
\includegraphics[width=\linewidth]{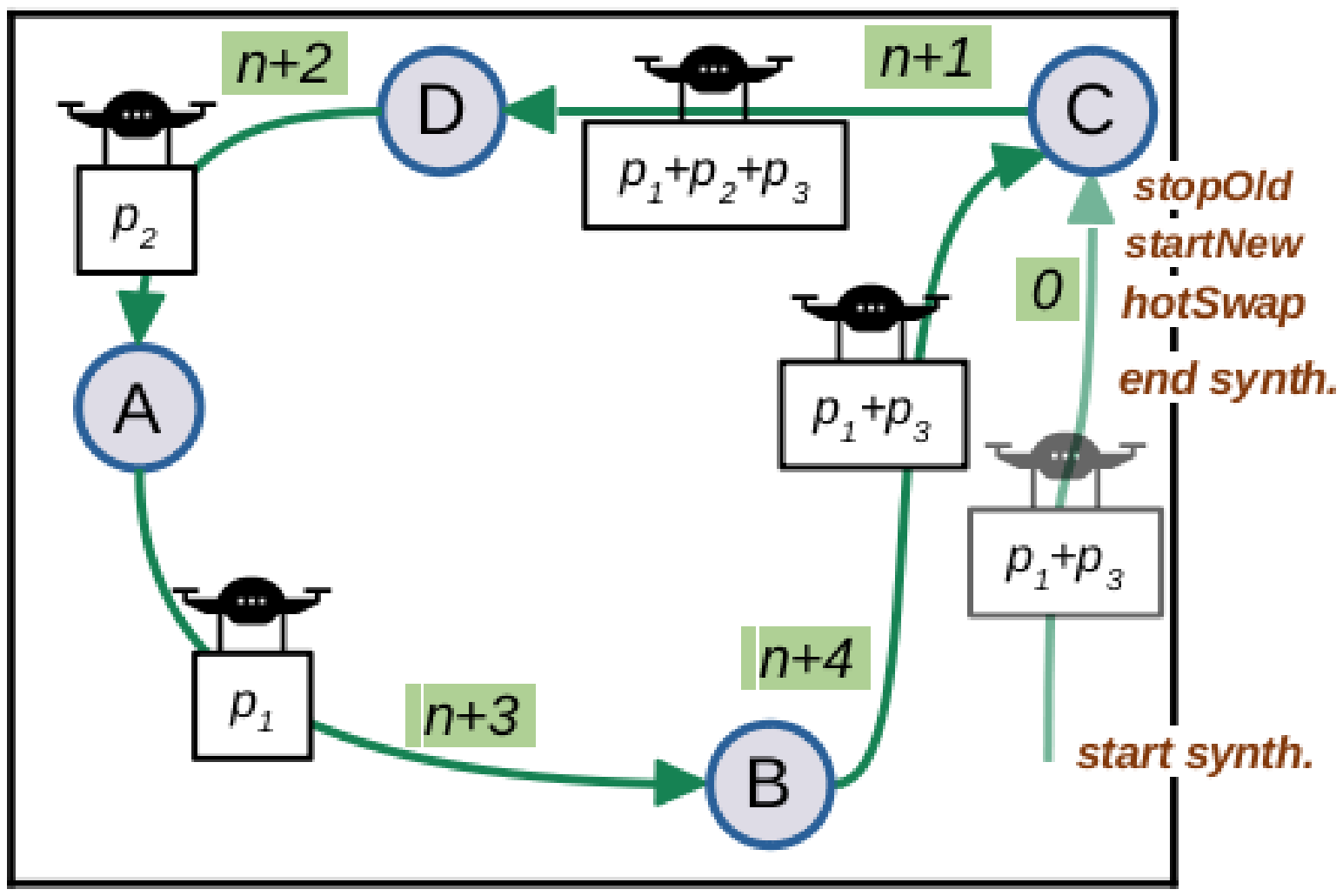}
\caption{Update plan when the \hotSwap occurs between \region{B} and \region{C}, while 
carrying $p_1$ and $p_3$.}
\label{fig:delivery2sol1}
\end{subfigure}
~
\begin{subfigure} {0.32\linewidth}
\centering
\vspace{3mm}
\includegraphics[width=\linewidth]{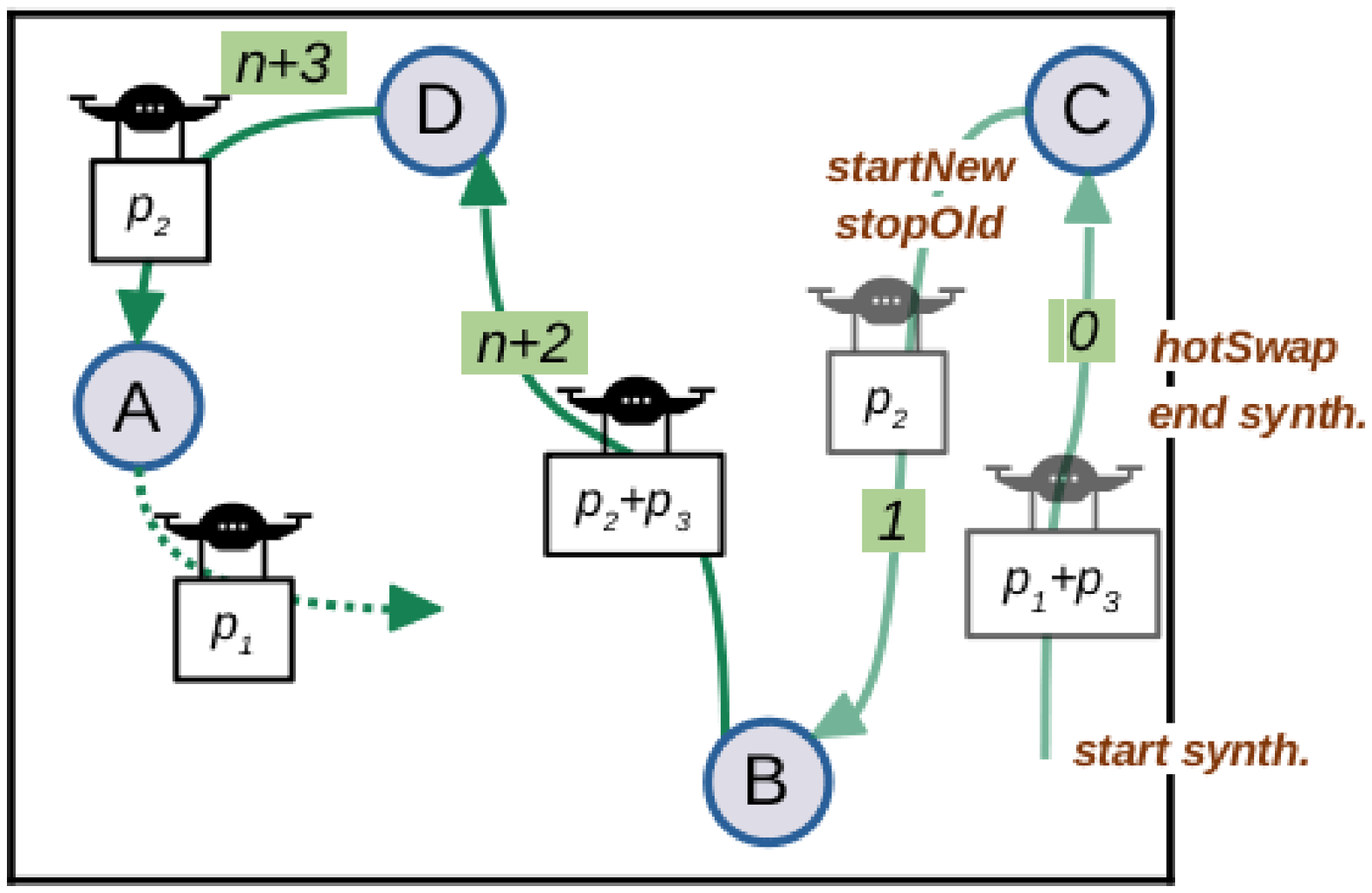}
\caption{Partial view of the update plan with weight requirements, showing delayed 
update when \hotSwap occurs in the same position as (d).}
\label{fig:delivery2sol3}
\end{subfigure}

\caption{Package delivery. Plans in (b), (d-f) 
are shown schematically without the discretized regions. Red arrows indicate 
required source and target of package type delivery. Green arrows are planned legs, 
labels indicate the order in which they occur ($n = 0\ldots$ indicates loop number). We 
omitted the occurrence of the event \reconfigure since in scenarios (d-f) reconfiguration 
is trivial.}
\label{fig:scenarios}
\end{figure*}

\emph{Example 1 (Delivery Service):} 
Consider a UAV
operating as a delivery messenger 
between three discrete locations \region{A}, \region{B}, and \region{C}, transporting 
package types 
$p_1$, 
$p_2$ and 
$p_3$ between them In Fig.\ref{fig:delivery1} we depict the pick-up and delivery 
requirements for each package type.  Additionally, it is required that the UAV must 
not move between locations without a package to preserve a minimum weight 
requirement. Assume the UAV is 
executing a plan depicted in Fig.~\ref{fig:delivery1sol}  that satisfies these 
requirements, travelling between \region{A}, 
\region{B}, and \region{C}, in that order, moving packages.  

Assume that at some point, while the UAV is flying, the mission 
needs to be updated to incorporate a new location \region{D} and different delivery 
requirements as depicted in Fig.~\ref{fig:delivery2}. Note that the location of the UAV 
is marked as unknown in Fig.~\ref{fig:delivery2} as the UAV is constantly moving 
while the requirements are 
being defined (and eventually deployed). The requirement of non-empty flights is 
maintained. 

\begin{figure}
\begin{subfigure} {0.6\linewidth}
\centering
\input{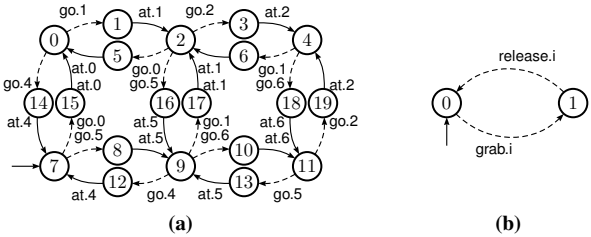}
\caption{}
\label{fig:discretization}
\end{subfigure}
~
\begin{subfigure} {0.3\linewidth}
\centering
\input{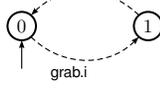}
\caption{}
\label{fig:package}
\end{subfigure}
\caption{(a) Movement restrictions in the discretized workspace (snippet). (b) 
Grab and release model for package $p_i$, with $i=1,2,3$.}
\label{fig:LTS}
\end{figure}

The original mission plan can be synthesised by defining a discrete abstraction 
for the workspace of the robot and constraining robot movements to adjacent 
cells. 
In Fig.~\ref{fig:discretization} we show a portion of the LTS covering only cells $0-2, 
4-6$, 
where we model the movement actions as control modes~\cite{ControlModes} with a 
controllable (\goN{i}) and 
uncontrollable (\arrivedN{i}) pair. We also model the grab and release mechanisms for the 
three package types $p_1, p_2, p_3$ as controllable actions using the LTS in 
Fig.~\ref{fig:package}.  Note that the initial location of the UAV is modelled by the 
initial 
state of the LTS of Fig.~\ref{fig:discretization}.

We now formalise the mission goals, specifying where 
each package type can be grabbed and released. For instance, we define a  safety 
property 
$\varphi_{p_{1}} = \Gl \big( (\grab{1} \implies \arrivedfl{4}) \wedge (\release{1} 
\implies 
\arrivedfl{3}) \big)$ to require packages of type $p_1$ be taken from \region{A} to 
\region{C}. 
Fluents 
$\arrivedfl{i} = \langle \set{\arrivedN{i}}, \set{\goN{k}\cdot 0 \leq k \leq 11}, \bot 
\rangle$ are true when the robot is at location $i$. Additionally we require  
$\psi_{p_1}  = \Gl \big(\with{1} \wedge \arrivedfl{3} \implies (\neg 
\formatFluent{Moving} \W \release{1}) \big)$ to ensure that the UAV will deposit packets 
as soon as it arrives at the respective target location, with fluents 
\formatFluent{Moving} and \with{i} being turned on/off with 
\formatAction{go}/\formatAction{at} actions and when the UAV does a 
\formatAction{grab}/\formatAction{release} of package $p_i$, respectively.
Avoiding empty trips is accomplished adding another safety 
specification: $\gamma = \Gl 
(\formatFluent{Moving} 
\implies \with{1} \vee \with{2} \vee \with{3})$.

Finally, we add the liveness property of continuously delivering packages 
$p_1,p_2,p_3$: $\rho = \Gl \F \release{1} \wedge \Gl \F \release{2} \wedge \Gl \F 
\release{3}$. 
We  refer to $\rho \wedge \gamma \wedge \bigwedge_{i:1, \ldots, 3} \varphi_{p_{i}} 
\wedge 
\psi_{p_{i}}$ as \old. 

A controller $C$ for this mission can be automatically built (as discussed in 
Sec.\ref{pre:CS}) by providing the specification (\old) and an environment $E$ (the 
parallel composition of the LTS in Fig.~\ref{fig:LTS}).  Note that 
\old can be rewritten as a combination of safety properties and a GR(1) property. The 
resulting controller (using MTSA) exhibits the 
following trace: \grab{1}, \goN{8}, \arrivedN{8}, \goN{9}, \arrivedN{9}, \goN{10}, 
\arrivedN{10}, \grab{3},  
\goN{11}, \arrivedN{11}, \goN{7}, \arrivedN{7}, \goN{3}, \arrivedN{3},  \release{3},  
\release{1}, \grab{2}, \goN{2}, ... .
A graphical depiction of the UAV being controlled is given in 
Fig.~\ref{fig:delivery1sol}.  

We now discuss adapting the mission plan to achieve delivery requirements of 
Fig.~\ref{fig:delivery2}. Note that there is no change in the discretization of the 
workspace and the functioning of the grab/release actuation modes. Thus, we 
can reuse the LTS models of Fig.~\ref{fig:LTS}  to define the environment model for 
the new mission plan (i.e., $E = E'$). The FLTL properties $\varphi_{p_{i}}$ and 
$\psi_{p_{i}}$  must be changed slightly to reflect the new delivery relations 
shown in Fig.~\ref{fig:delivery2}. Assume these properties to be $\varphi_{p_{i}}'$ and 
$\psi_{p_{i}}'$.  We refer to $\rho \wedge \gamma \wedge \bigwedge_{i:1, \ldots, 3} 
\varphi_{p_{i}}' 
\wedge  \psi_{p_{i}}'$ as \new.

To perform a mission update we must formulate a DCU problem. which requires not 
only the old and new mission specifications but also two further inputs: A function 
$g$ 
from $E$ states to $E'$ states is required and a transition property $\Theta$ 
constraining (if needed) the occurrence of \stopOld, \startNew, and 
\reconfigure.

Given that $E = E'$, we define $g$ as the identity function. This means that the new 
controller when in place will assume that the current state of the environment $E'$ is 
the same as the current environment of $E$. Or more precisely, at the occurrence of 
the \reconfigure event, the execution environment of the controller can be assumed 
to behave as $E'$ setting its initial state based on the current state of $E$. 

For this simple example, it suffices to used the \textit{standard 
transition requirement}  $\Theta_\emptyset = \Gl (\neg \stopOldfl \vee \startNewfl)$ 
mentioned in Section~\ref{sec:prelimDCU} requiring the system to be satisfying one of the 
two 
mission requirements. We will discuss more complex transition requirements in the 
next examples. 

Having defined $\new$, $E'$, $g$ and $\Theta$, and given that the current mobile robot is 
running a controller $C$ to achieve $\old$ in environment $E$, we have a fully formulated 
DCU problem (with live missions, see Section~\ref{sec:prelimDCULive}) for which a 
solution ($C$ and $f$) 
can be constructed.

Consider the scenarios in Fig.~\ref{fig:delivery2sol2} and 
\ref{fig:delivery2sol1} in which the UAV controlled by $C$ is flying towards location 
\region{C} carrying two packages as part of its plan to satisfy $\old$ when a synthesis 
procedure to find a solution to the DCU problem is run. The scenarios differ in when the 
synthesis procedure ends (see \emph{end synth.})

In Fig.~\ref{fig:delivery2sol2}, while $C'$ is being computed the UAV reaches location 
\region{C} and, as per $\old$ drops off package $p_1$ and $p_3$, picks up $p_2$ and then 
continues 
towards location \region{A} where it must deliver $p_2$. On flight towards \region{A}, 
the DCU synthesis procedure ends, the new controller $C'$ to be hotswapped in and its 
current state is set in terms of the current state of $C$ and function $f$. At 
this point $C'$ 
declares declares that $\new$ will hold from now on ($\startNew$) and that $\old$ will not 
be guaranteed anymore ($\stopOld$). It does so in this order to comply to $\Theta$. 
Function $f$ has been computed to preserve the state of $C$ in $C'$, thus $C'$ ``knows'' 
that it is carrying $p_2$ and is on its way to \region{A}. From then on $C'$ commands the 
UAV as per $\new$.

In Fig.~\ref{fig:delivery2sol1}, $C'$ and $f$ are computed before reaching \region{C}. 
This time controller $C'$ is  hotswapped in before the UAV reaches \region{C} and its 
initial state is set differently by $f$ than before because $C$ is in a different state. 
Now $C'$ ``knows'' that it is carrying $p_1$ and $p_3$ and is on its way to \region{C}. 
The new controller declares $\startNew$ and $\stopOld$ and upon reaching \region{C} it 
can no longer (as in the previous scenario) drop $p_1$ and $p_3$ as it would be 
inconsistent with $\new$. Instead, it picks up $p_2$ and continues pickups and drop offs 
as per $\new$.

Note that when the computation of $C'$ and $f$ started, no assumption is made as to 
whether the computation will end before or after the UAV reaches location \region{C}. 
Thus, it is the same $C'$ and $f$ that are computed in both scenarios. This demonstrates 
the need for $C'$ to have a strategy for transitioning into the new mission that works for 
any state in which the current system might be in. It is $f$ at how-swapping time 
determines which state $C'$ should be set at, and consequently which transition strategy 
should be used.

In the two previous scenarios, the mission switch was performed immediately after 
hotswapping controllers. This is not always the case. 
Consider a scenario in which the user introduces into the $\new$ an additional 
requirement forbidding  transportation of three packages (to avoid overstraining the 
UAV): $\Gl \neg (\with{1} \wedge \with{2} 
\wedge \with{3})$. Assume that similarly to Fig.~\ref{fig:delivery2sol1} the computation of 
$C'$ and $f$ terminates before the UAV reaches location \region{C} (see 
Fig.~\ref{fig:delivery2sol3}). Here, the new controller cannot immediately start satisfying 
$\new$ as picking up $p_2$ would violate the requirements. Hence, it chooses to 
\textit{delay} the change of specification, first dropping off $p_1$ and $p_3$, and also 
picking up $p_2$ as required in $\old$. Only then, it switches mission and flies to 
\region{B} 
rather than \region{A}.

Note that we have deliberately omitted referring to the occurrence of event $\reconfigure$
for simplicity. We discuss this event in subsequent scenarios.

\subsection{Dealing with Re-Discretization and new Capabilities}

\begin{figure*}
\begin{subfigure} {0.23\linewidth}
\centering
\includegraphics[width=\linewidth]{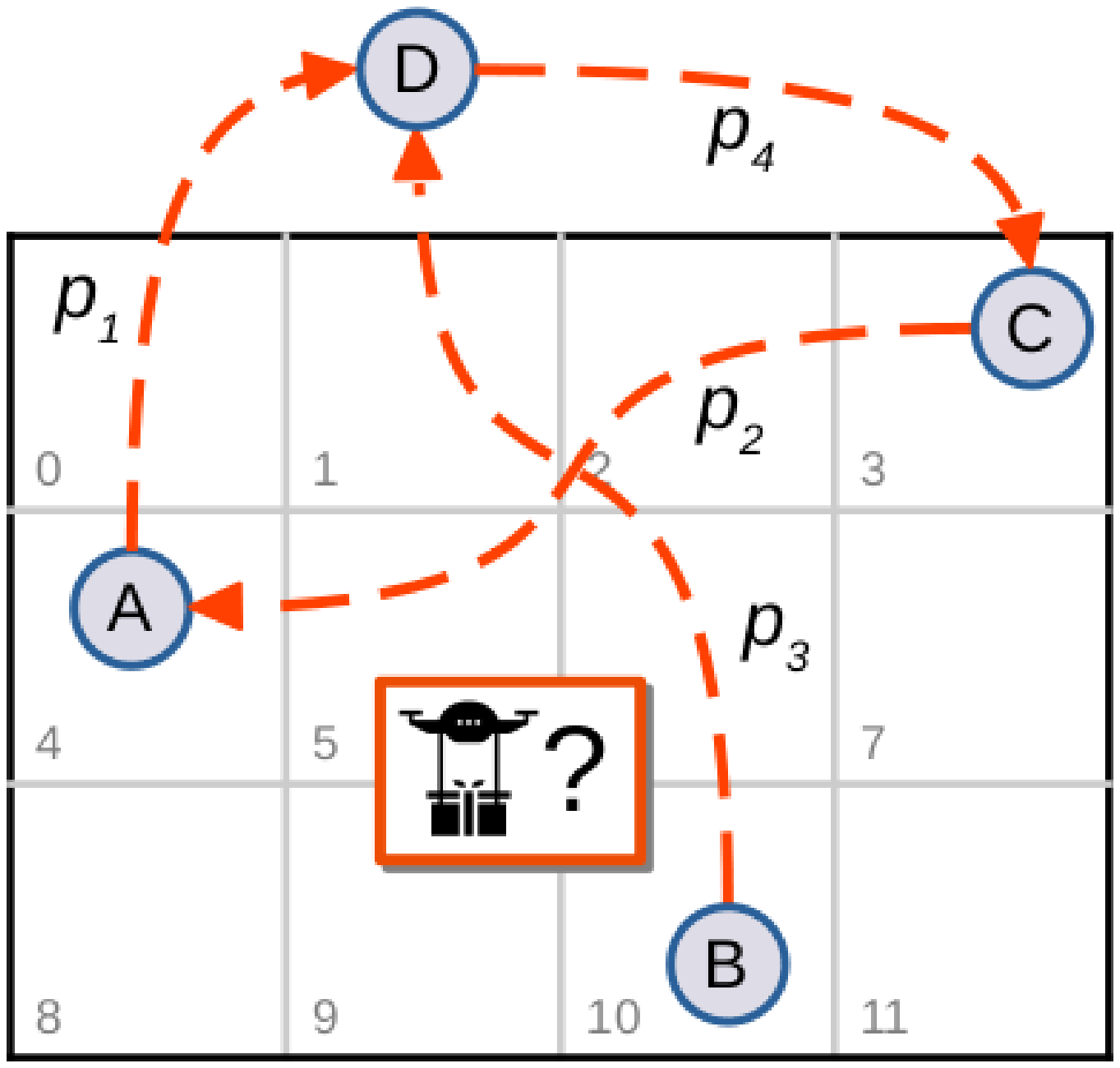}
\caption{}
\label{fig:delivery3}
\end{subfigure}
~
\begin{subfigure} {0.14\linewidth}
\centering
\input{src/lts/packetreconf}
\caption{}
\label{fig:reconf-packet}
\end{subfigure}
~
\begin{subfigure} {0.23\linewidth}
\centering
\includegraphics[width=\linewidth]{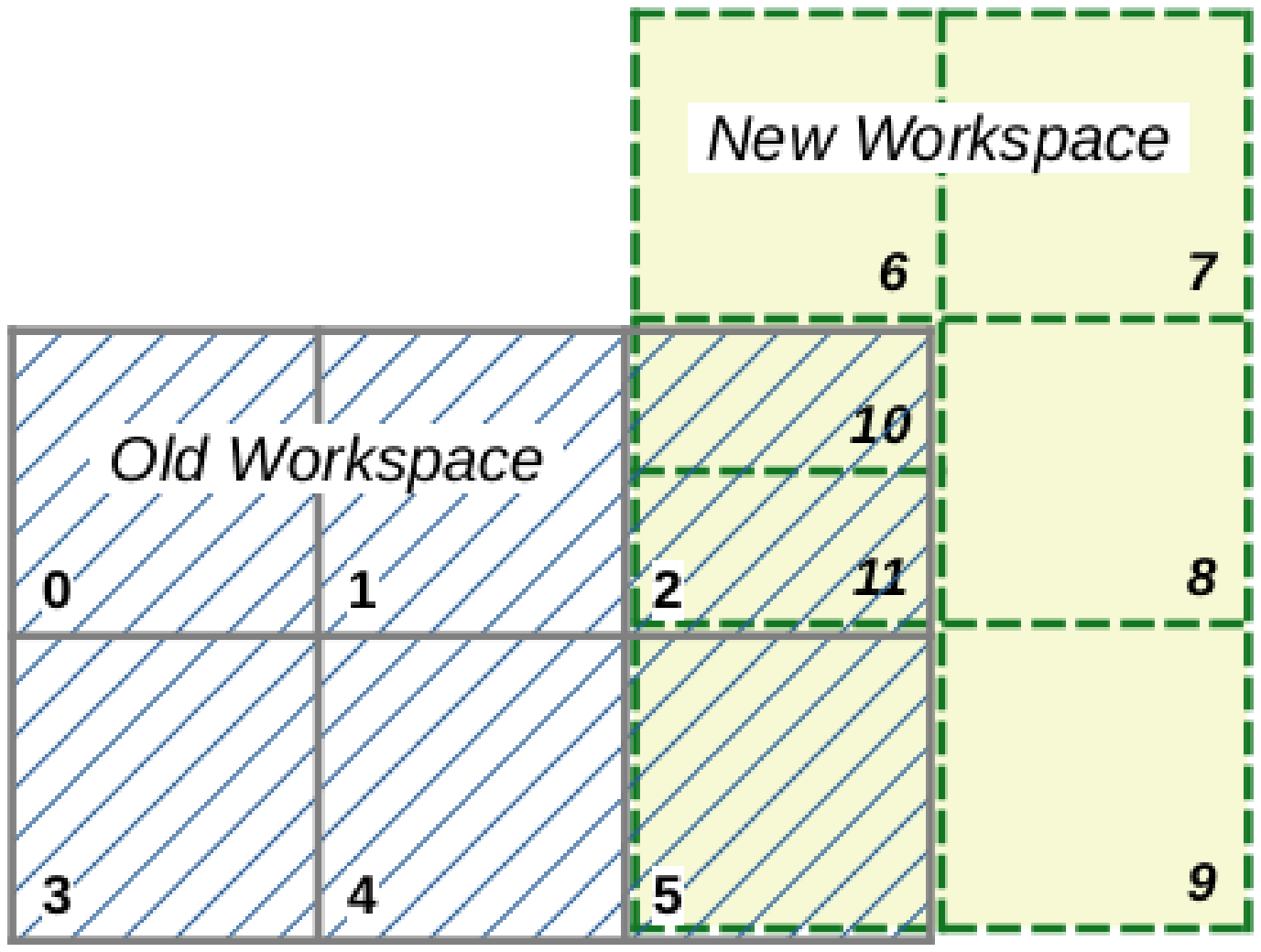}
\caption{}
\label{fig:reconf-work}
\end{subfigure}
~
\begin{subfigure} {0.37\linewidth}
\centering
\input{src/lts/discretizationChangesplit}
\caption{}
\label{fig:reconf-map}
\end{subfigure}
\caption{(a) Variant of updated delivery mission. (b) 
Reconfiguration model for packet $p_4$. (c) Simplified scenario of workspace 
reconfiguration. (d) Reconfiguration model for the workspace in (c), with all the 
\arrivedN{i} transitions removed for clarity.}
\label{fig:packetsupd}
\end{figure*}

The simple example from the previous section avoided a key difficulty in real mission 
adaptation: \textit{what happens when the new mission requires o must deal with a 
change in the 
execution environment of the robot?} By execution environment we refer to hardware 
that may be malfunctioning,  software with new sensor or actuating capabilities that 
must 
be uploaded, or changes in the assumptions that are considered valid given the 
conditions of the physical world in which the robot is operating. Ultimately, from a 
control perspective all these changes represent changes in the set of controllable 
and non-controllable events, the formulation of the environment LTS $E'$ and mission 
goals $\varphi'$.

Consider the following example:

\emph{Example 2 (Reconfiguring Delivery Service):} Assume the original mission 
specification from \textit{Example 1}. A new pick-up/drop-off location \region{D} must be 
introduced. The location falls beyond the current discretized workspace. In addition,  a 
new package type $p_4$ is to be transported. The delivery requirements 
are shown in Fig.~\ref{fig:delivery3}. The non-empty trips requirement is kept.
The distinctive shape of package type $p_4$ requires a software module 
tailored specifically to control the robot's 
gripper to successfully pick them up and drop them off.

The DCU control problem uses the $\reconfigure$ to model the change in the 
execution environment of the robot. For this example, two reconfiguration aspects 
need to be modelled. First, a model for the $p_4$ grab release module must be 
introduced, ensuring that its initial state is one in which no $p_4$ package is being 
held (see Figure~\ref{fig:reconf-packet}). The second, is the reconfiguration of the 
workspace discretization,
for which we use the simplified workspace change depicted in 
Fig.~\ref{fig:reconf-work}. 
Here some of the discrete cells are only present in the old workspace (0, 1, 3, and 
4), some are only present in the new workspace (6--9), cell 5 is present in both and 
there is a change of granularity for cell 2, which now maps to 10 and 11. We 
model the mapping between the states of the old and new environments in  
Fig.~\ref{fig:reconf-map}. Note that reconfiguration
can only happen when the robot is in one of the shared discrete cells (2 and 5), where 
the choice of where cell 2 maps is non-deterministic from the controllers perspective.

Composing the LTS in Fig.~\ref{fig:reconf-packet} for $p_4$, with the models for $p_1, 
\ldots, p_3$ (see Fig.~\ref{fig:package}) and one along the lines of 
Fig.~\ref{fig:reconf-map}, generates a model for $E\lig{g}{\reconfigure} E'$ (instead of 
providing 
$E'$ and $g$ separately). 

The new delivery requirements are modelled according to Fig.\ref{fig:delivery3} must be  
included in $\new$, and a transition requirement $\Theta$ must be provided. We 
assume the simple $\Theta_\emptyset$ used previously that requires the UAV to always be 
constrained according to either of the missions. 

\begin{figure}
\begin{subfigure} {0.6\linewidth}
\centering
\includegraphics[width=\linewidth]{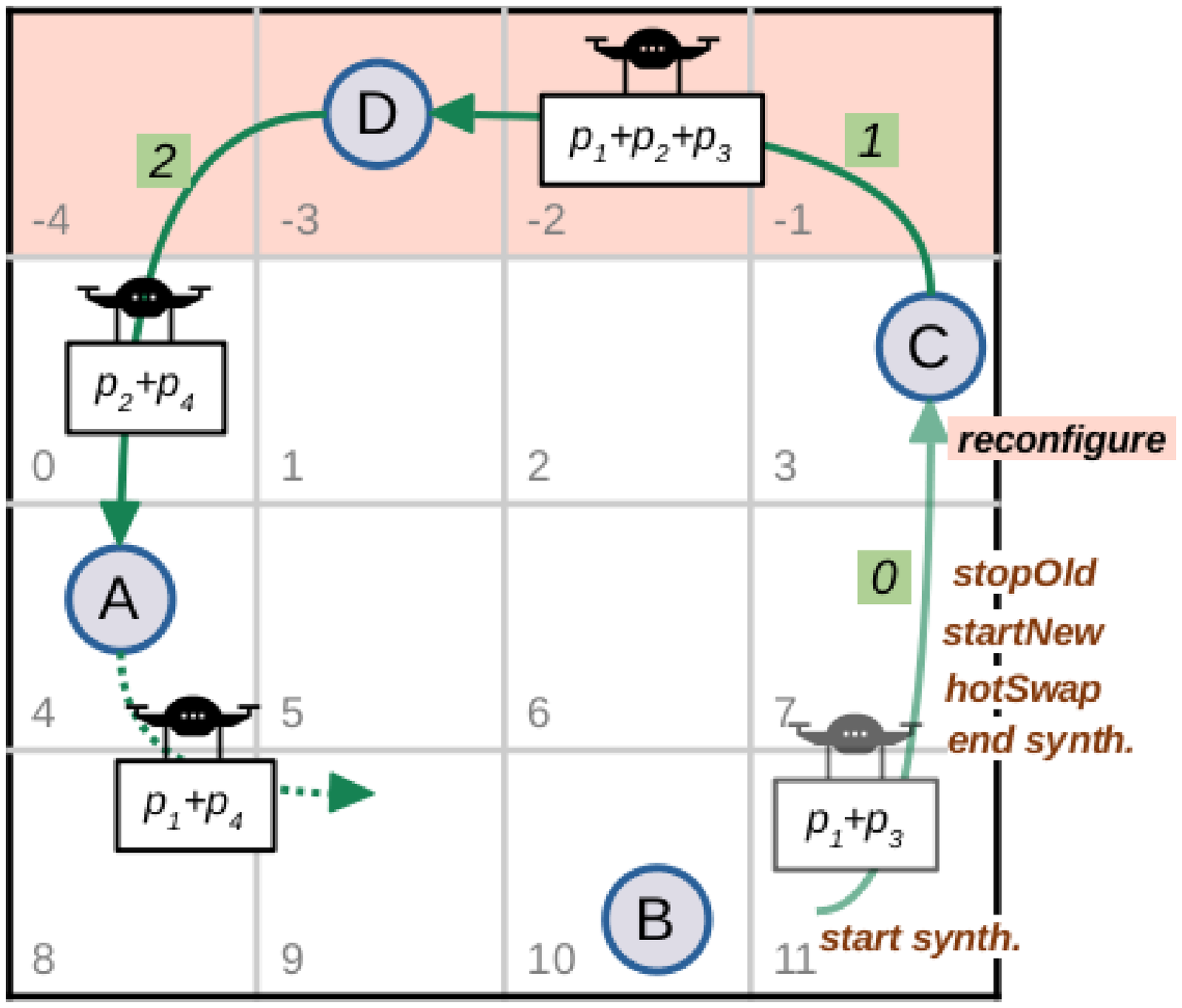}
\caption{}
\label{fig:delivery3sol}
\end{subfigure}
~
\begin{subfigure} {0.37\linewidth}
\centering
\includegraphics[width=0.8\linewidth]{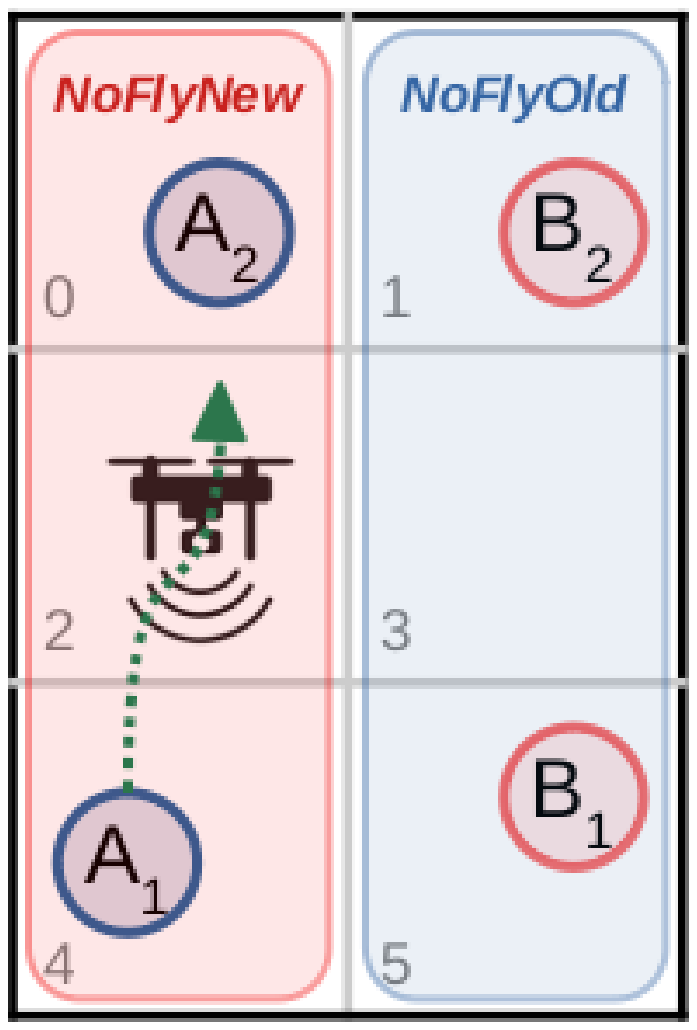}
\caption{}
\label{fig:patrol}
\end{subfigure}
\caption{(a) Partial view of the update plan for the scenario in 
Fig.~\ref{fig:delivery3}, where \hotSwap occurs between $B$ and $C$. (b) Inconsistent 
patrol mission update.}
\end{figure}

The resulting DCU problem can be solved and a new controller 
$C'$ and controller initialization function $f$ can be computed for the \textit{Example 
2} scenario. We show in Fig.~\ref{fig:delivery3sol} an update scenario that the new 
controller may 
exhibit. Synthesis starts and ends with the UAV on its way to location \region{C}. The 
new 
controller $C'$ is hotswapped in and upon arriving to location 3 (\arrivedN{3}), the 
controller commands a reconfiguration. This is possible because location $3$ is part of 
the old and new discretized workspace. With $\reconfigure$, the UAV infrastructure is 
changed: a new module for grabbing and releasing $p_4$ packages is added to the 
software architecture. The UAV is then commanded to location \region{D} via newly 
introduced discrete locations ($-1,...,-4$). At $-3$, the new module is used to pick up 
a $p_4$ package. Control proceeds satisfying the new 
mission requirements.

\subsection{Inconsistent Mission Adaptations} \label{subsec:trans}
Sometimes, the behaviour of a new mission may be logically inconsistent with the 
current UAV mission. For these cases, the transition requirement $\Theta_\emptyset = \Gl 
(\neg \stopOldfl \vee \startNewfl)$ which we used in the previous scenarios is not 
adequate: If the two 
missions are logically inconsistent, there is no safe state in which to first do 
$\startNew$ and then $\stopOld$. We illustrate this in a simple example. 
%
%

\emph{Example 3 (Surveillance Update):} Consider a typical UAV patrol mission as described 
in~\cite{Menghi19} for surveillance of two areas $A_1$ and $A_2$ as shown in 
Fig.~\ref{fig:patrol}. To restrict the movement
area of the UAV the user imposes an area \region{\nfold} as
a no-fly zone allowing other vehicles or humans to work
in this region. The user now decides that the surveillance must now be done between areas 
$B_1$ and $B_2$, and moves the no-fly zone to \region{\nfnew}, as shown in 
Fig.~\ref{fig:patrol}.

The original UAV mission goal can be written as:
$ \left(\Gl \F \arrivedfl{0} \wedge \Gl \F \arrivedfl{4} \right)$ $\wedge$ $\left( \Gl 
 \neg \arrivedfl{\region{\nfold}} \right)$.  That is, be at cells $0$ and $4$ infinitely 
often and never be within the \region{\nfold} region. Similarly,  the new 
mission can be specified
as $\left(\Gl \F \arrivedfl{3} \wedge \Gl \F 
\arrivedfl{5} \right) \wedge \left( \Gl \neg 
\arrivedfl{\region{\nfnew}} \right)$, using appropriately defined fluents.

Note that in this example if we used the transition requirement $\Theta_\emptyset$
the DCU problem has no solution, as is not possible switch from the old goal 
(\old) to the new goal (\new) without violating one of them. If the UAV is in 
locations from the left column ($0,2,4$) then it cannot switch to achieving \new 
 because these locations are in the new no-fly area. If the UAV is moved to locations 
from the right column to comply to \new then it is violating \old. This is 
an extreme example that motivates the need for specifying requirements that deal with 
\textit{transitioning behaviour} between missions. 

A trivial but unsatisfactory solution to resolving inconsistencies is to impose no 
transition requirements $\Theta = \top$. However, this allows arbitrary behaviour: the 
controller may declare $\stopOld$, and once relieved of following the old mission 
requirements perform arbitrary actions before performing $\startNew$. Note that the 
latter must eventually occur as $\F \startNew$ is required. 

To resolve the inconsistency between the old and new no-fly zones 
in the  patrol mission change, a reasonable transition requirement may be to allow a 
period in which neither old nor new mission restrictions are satisfied but restrict what 
can occur during this period. For instance, to restrict movement between no-fly zones 
to the bottom side of the grid. That 
is, when the old specification is dropped, the UAV must be at locations $4$ 
or $5$ until the new one is adopted: 
\begin{equation} \label{eqn:T-arrived45}
\Theta = \Gl \Big(\stopOldfl \implies
\big((\arrivedfl{4} \vee \arrivedfl{5}) \W \startNewfl \big) \Big)
\end{equation}

With this transition requirement, it is possible to synthesise a mission to satisfies the 
mission adaptation. 


%
%
%
%


\section{Adaptive Architecture and Implementation} \label{sec:morphhybrid}

\begin{figure*}
\begin{subfigure} {0.45\linewidth}
\centering
\includegraphics[width=\linewidth]{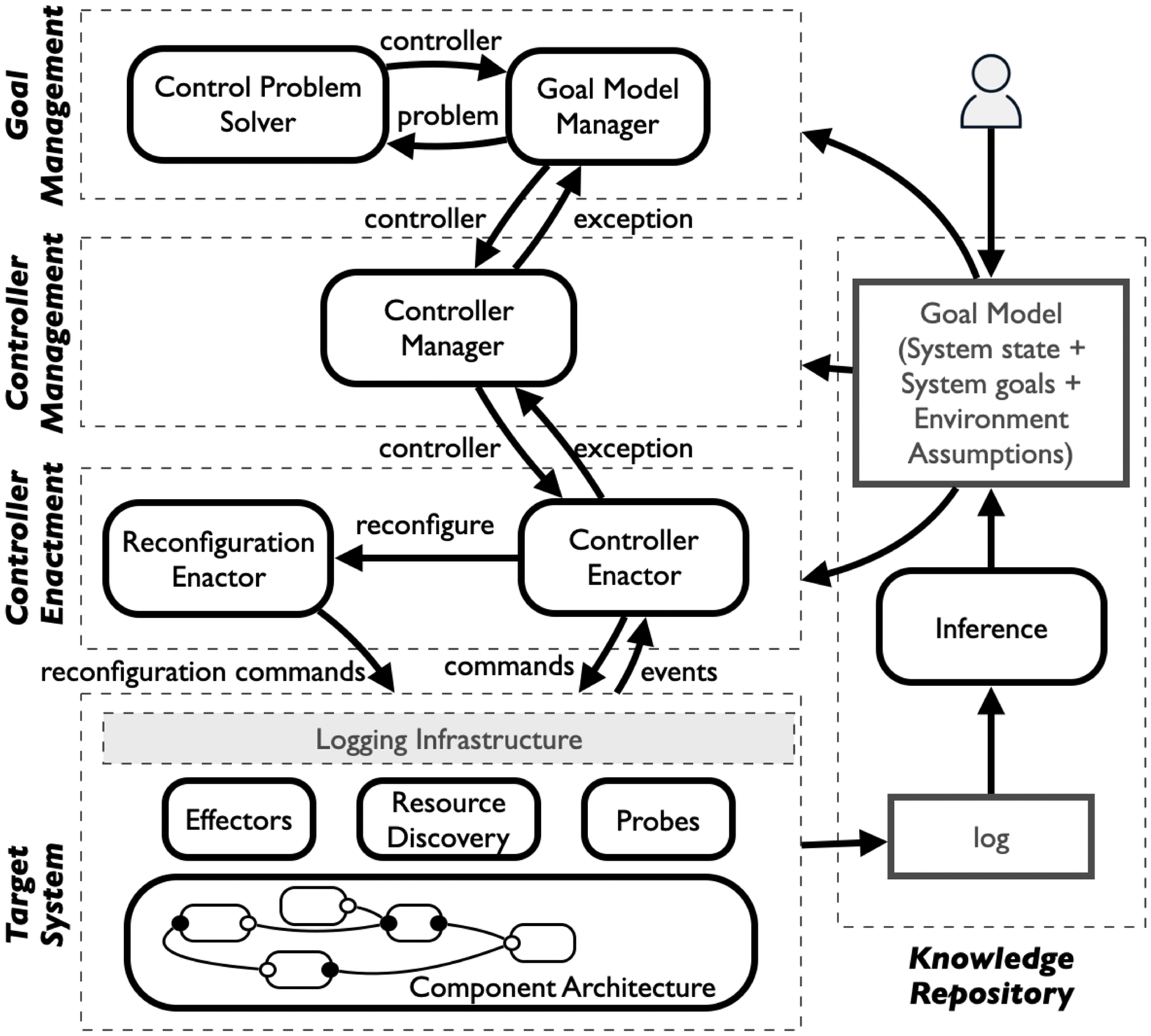}
\caption{MORPH Reference Architecture}
\label{fig:morpharch}
\end{subfigure}
~
\begin{subfigure} {0.55\linewidth}
\centering
\includegraphics[width=\linewidth]{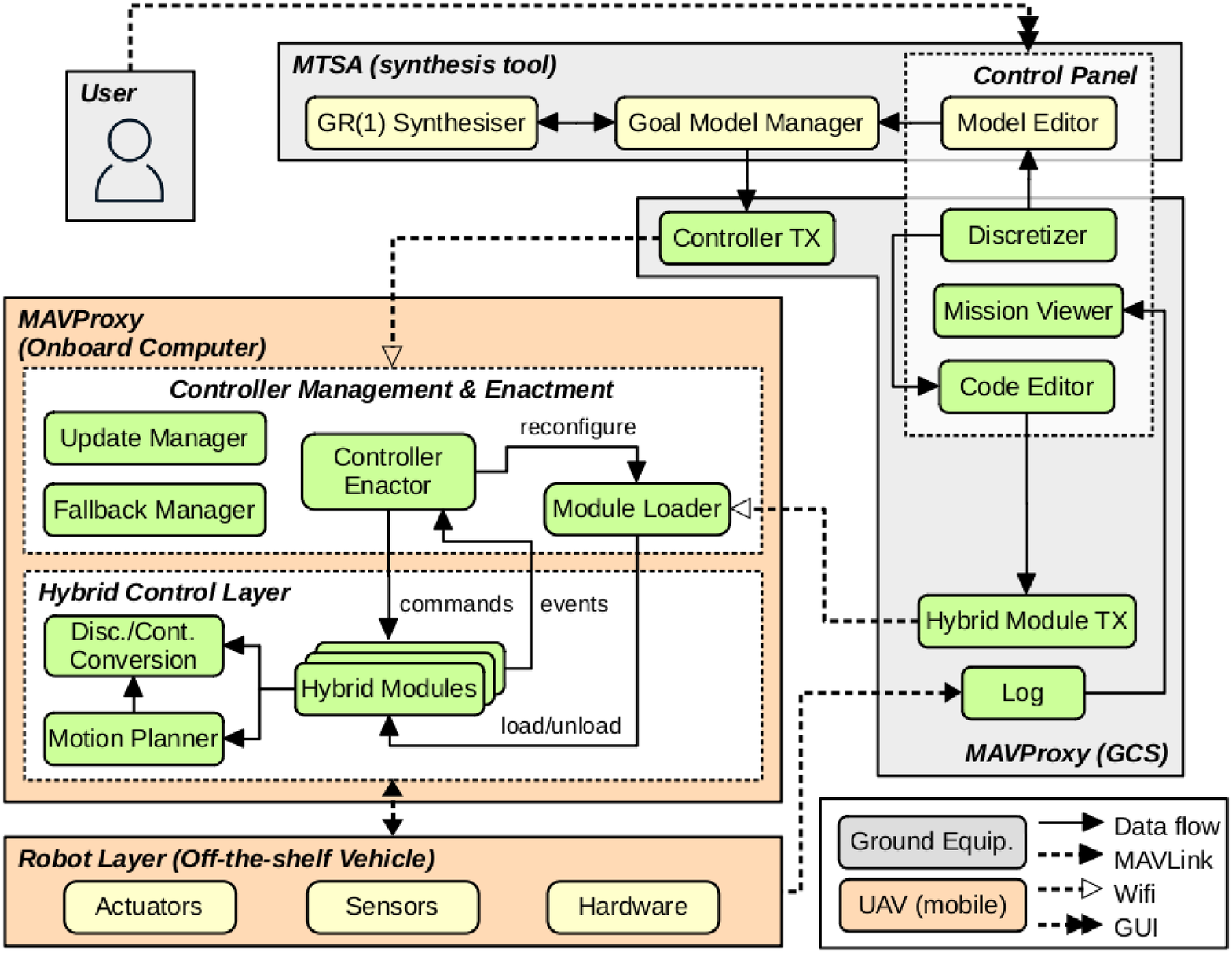}
\caption{Architecture for Assured Mission Adaptation of UAVs}
\label{fig:dynamicarch}
\end{subfigure}
\caption{Adaptive Architectures}
\end{figure*}

In the previous section we showed how to cast assured mission adaptation as a DCU 
problem. In this section we will show how a solution to a DCU problem can be used 
in a robotic system to effectively provide assured mission adaptation. To this end, we 
build on the notion of hybrid controller~\cite{Gazit09}  to address three 
implementation challenges: 
(a) 
\textbf{uploading and hotswapping}  
new discrete controllers at runtime, (b) loading and unloading of software 
components to allow \textbf{coordinated software reconfiguration}, (c) 
\textbf{human-in-the-loop support} for runtime specification of mission 
adaptations.
We address these challenges by taking elements from the MORPH~\cite{MORPH15} 
reference architecture and integrating them with hybrid controllers. MORPH  
(Figure~\ref{fig:morpharch})  outlines a framework for architectural adaptation 
through the runtime synthesis, hotswapping and enactment of 
correct-by-construction strategies. 

Firstly, we explain how MORPH and hybrid architectures resolve the three 
implementation challenges while providing links to the concrete architecture we 
implemented (Figure~\ref{fig:dynamicarch}). Secondly, we  provide implementation 
details of the architecture that supports assured mission adaptation of UAVs.



\subsection{Architectural Extensions for Adaptability} \label{sec:extension}

Hybrid controllers serve as an interface between the discrete high-level events 
of the synthesised controllers  and the low-level sensors, feedback-controllers and 
other actuators from a robot. As a result, 
these architectures help  produce continuous movement and trajectories that 
satisfy the user specification. In terms of MORPH, this hybrid layer lives in the  
\textit{Target System}, while the continuous execution of the discrete event 
controller is what occurs within the \textit{Controller Enactor} component (see 
Figure~\ref{fig:morpharch})
The MORPH \textit{Target System} is implemented in our architecture by the  
\textit{Robot Layer} and the \textit{Hybrid Control Layer} (see 
Figure~\ref{fig:dynamicarch}). The MORPH \textit{Effectors} map 
to \textit{Actuators} and \textit{Actuators}, while \textit{Probes} 
map to \textit{Sensors}, together with the required \textit{Hybrid Modules} to make 
them work.

MORPH proposes dealing with controller \textbf{uploading and hotswapping} as 
follows: The \textit{Controller Management Layer} is responsible for  
replacing the controller currently being enacted by a new one. This can be triggered 
by the reception of a new controller from the \textit{Goal Management Layer} or due 
to an exception raised by the \textit{Controller Enactor} (the \textit{Controller 
Management Layer} may store fallback controllers). Note that MORPH does not 
provide any guidance or mechanisms to ensure that the hotswapping is correct, 
neither does it define correctness for that matter. 

In our implementation, the MORPH \textit{Controller Management Layer} is a component 
(\textit{Update Manager}) that uses the output of MTSA synthesis tool for a DCU 
control problem to hotswap the current controller in the \textit{Controller Enactor}. 
To do so, it must take the MTSA output, $C\lig{f}{\hotSwap} C'$ where $C$ is the 
controller currently running in the \textit{Controller Enactor}, extract $C'$ and $f$, 
identify the current state of $C$, hotswap $C$ with $C'$ within the \textit{Controller 
Enactor} and set the state of $C'$ according to $f$. It must do this procedure 
atomically. 

Our implementation also includes a \textit{Fallback Manager} that provides a preset 
fallback discrete event controller that is to be used if an event is received that is not 
enabled in the current state of the  controller being enacted. 

The MORPH \textit{Goal Management} layer's responsibility is to to produce 
controllers for the \textit{Controller Management} layer. It constructs control problems 
based on a \textit{Knowledge Repository} and uses a \textit{Control Problem Solver} to 
produce discrete event controllers. In our implementation, the MTSA tool (see top of 
Fig.~\ref{fig:dynamicarch}) implements both the \textit{Goal Management} layer and 
part of the \textit{Knowledge Repository} by providing functionality for representing 
knowledge of the robots capabilities, environment assumptions and mission goals, a 
GR(1) synthesis procedure and transformation procedures for various control 
problems (including DCU) to \gr.  The implementation also includes  the  
\textit{Controller TX} module for uploading the result of MTSA to the robot.

In MORPH, \textbf{software reconfiguration} is also considered. Note that in 
Figure~\ref{fig:morpharch} a simplified version of software reconfiguration is 
depicted, one in which it is assumed to be atomic; this is not always the case. The 
\textit{Controller Enactor} commands a \textit{Reconfiguration Enactor} to reconfigure 
and the latter then reconfigures the \textit{Target System}.  How new software 
modules are loaded is unspecified in MORPH.

In the concrete architecture we developed, reconfigurations are limited to basically 
adding and removing hybrid modules that implement abstract events and commands 
that may appear in a discrete event controller (e.g., new 
capabilities, a different \textit{Motion Planner}) and modules for 
discrete to continuous conversion of discrete locations (i.e., allowing 
re-discretization).  New modules (and their mapping to events/actions) are received 
and stored by the \textit{Module Loader}. Instructions for unloading unnecessary 
modules can also be received. Upon reception of the \reconfigure command the 
\textit{Module Loader} loads and unloads the corresponding modules into the 
\textit{Hybrid Control Layer}.

MORPH prescribes \textbf{human-in-the-loop support} for adaptation via a 
\textit{Knowledge 
Repository}. This repository accumulates knowledge from logged data from the 
\textit{Target System} and combines it with user supplied input to close the 
adaptation 
loop. How this information is specified, inferred and stored is not prescribed by 
MORPH, however it is assumed that all elements required for deriving adaptation 
strategies are provided via the \textit{Knowledge 
Repository}.

Specifics of the tools used to implement the user interaction and the 
\textit{Knowledge 
Repository} are provided in the next section. The main components however are:

\begin{itemize}
 \item \textit{Model Editor:} Text editor to specify the original and update synthesis 
problems using LTS models and FLTL formulae.
 \item \textit{Discretizer:} Interactive map that allows the user to select discrete 
regions and their granularity to aid in the specification of the synthesis problems, 
automatically generating representations in Python code, LTS and FLTL that are used 
by 
the \textit{Model Editor} and \textit{Code Editor} components.
  \item \textit{Mission Viewer:} Collection of graphical representations of the data 
  from 
the \textit{Log} that provide the user significant input about the ongoing mission.
 \item \textit{Code Editor:} Standard code editor that allows the user to program the 
 new software modules required during the adaptation.
\end{itemize}

\subsection{Implementation on UAVs} \label{subsec:implementation}

We now discuss implementation specifics of an architecture for assured mission 
adaptation of UAVs, including the main architectural hardware components 
and a key robotics software package we used: MAVProxy. 

The system (see Fig.~\ref{fig:dynamicarch}) comprises of three main hardware 
components: An \morphOfftheshelf, an 
\morphComputer that runs the adaptation software, and a general 
purpose computer that acts as a \morphGCS (GCS). 

The \morphOfftheshelf includes hardware required for flying 
(propellers, rudders, batteries, motors, etc.) and an embedded processor that runs 
communications software supporting MAVLink~\cite{MAVLink}, a 
lightweight messaging protocol for receiving commands and sending 
telemetry. The processor also runs software for its sensors and actuators including 
various feedback-control loops (often referred as Autopilot or Flight 
Controllers) that implements MAVLink commands such as commanding the 
vehicle to navigate to a specified waypoint, calibrating sensors,  return-to-launch 
emergency commands, and arming and disarming the vehicle. 
In our experimentation we  used a Parrot Ar.Drone 2.0~\cite{Parrot,Parrot2}, a simple 
quadcopter, that has proprietary communication with basic MAVLink capabilities. We 
also used the ArduPilot Software-In-The-Loop (SITL) UAV simulator as 
in~\cite{Baidya18,Barros16}. The SITL simulator allows us to test UAV systems loaded with 
ArduPilot firmware (e.g., custom made as in~\cite{Rokhmana16} or commercial as the 3DR 
Solo Drone~\cite{Baidya18}) without the UAV hardware. This simulator 
(see~\cite{Iterator}) has been used
to seamlessly go from testing to actually flying a custom made 
fixed-wing vehicle (e.g.,~\cite{Mapping}) 
based on a Pixhawk (e.g.,~\cite{Sinisterra17,MAVProxy3}).

Similarly to~\cite{MAVProxy1, Yamazaki19, Rasp2} we expand the computing 
capabilities on the vehicle by physically fixing on top a general purpose \morphComputer
that runs most of the adaptive software architecture. We use a Raspberry Pi 3B+ single 
board computer when simulating with the ArduPilot SITL (to emulate realistic computing 
capabilities for an onboard computer), and a lighter Raspberry Pi Zero 
W mounted on the vehicle when flying the Parrot Ar.Drone 2.0.
  
Finally, for the \morphGCS, a standard laptop computer is 
needed to run the discrete event controller synthesis software \cite{MTSA}.
 We used a Linux laptop with an Intel i7 3.5GHz processor and 12GB of RAM.

A key element of the architecture is the MAVProxy software package~\cite{MAVProxyGit}. 
This is a widely used (e.g.,~\cite{MAVProxy3, MAVProxy1, MAVProxyGit, MAVProxy2}) GNU 
GPL Python 
  package for UAVs that implements the MAVLink protocol. The package includes 
many standard modules, from firmware management to a camera viewer and a 
moving map, that allow configuring a MAVProxy process to support different vehicle 
setups and mission tasks.

MAVProxy is designed 
to be used as ground control software (e.g.,\cite{MAVProxy3}). That is, MAVProxy 
running on a computer on the ground and providing a 
series of modules that provide mission monitoring capabilities and high-level commands 
for 
setting up and running missions. However, since MAVProxy provides a simple 
mechanism using custom modules for ad-hoc extensions, 
 it has also been used onboard the vehicle (e.g.,~\cite{MAVProxy1}) to provide new 
functionality.

The architecture runs a lean MAVProxy instance on the \morphComputer
including 
only default standard modules that connect via WIFI or serial communication 
with the embedded processor on the \morphOfftheshelf. However, we 
add a number of custom modules to implement \textit{Controller Managemente \& Enactment}, 
and \morphHybridControl layers. 

Note that the system conformed by the \morphComputer running the 
MAVProxy instance and the \morphVehicle running the flight controllers is a fully 
autonomous vehicle that does not need communication to a ground computer to 
fly its mission.

The \morphGCS runs a separate MAVProxy instance, 
configured similarly to most MAVProxy uses. It communicates 
directly with the vehicle to perform the initial mission setup, to receive telemetry and 
to allow taking control over the vehicle if necessary. Telemetry data is shown on a GUI 
to users using standard modules. This ground MAVProxy instance communicates 
with the airborne MAVProxy instance using a custom protocol over TCP/IP via WIFI to 
support the interactions between the \morphCtrlManagment layer and 
the \morphSynthesiser. The \textit{Control Panel} is conformed with joint elements from 
MAVProxy and MTSA, which provides functionality for specifying discrete event control 
problems as described in Section~\ref{sec:Preliminaries} and provides a back-end 
\morphSynthesiser layer.

\section{Validation}
\label{sec:Validation}
\begin{figure*}
\begin{subfigure} {0.335\linewidth}
 \centering
\includegraphics[width=1.0\linewidth]{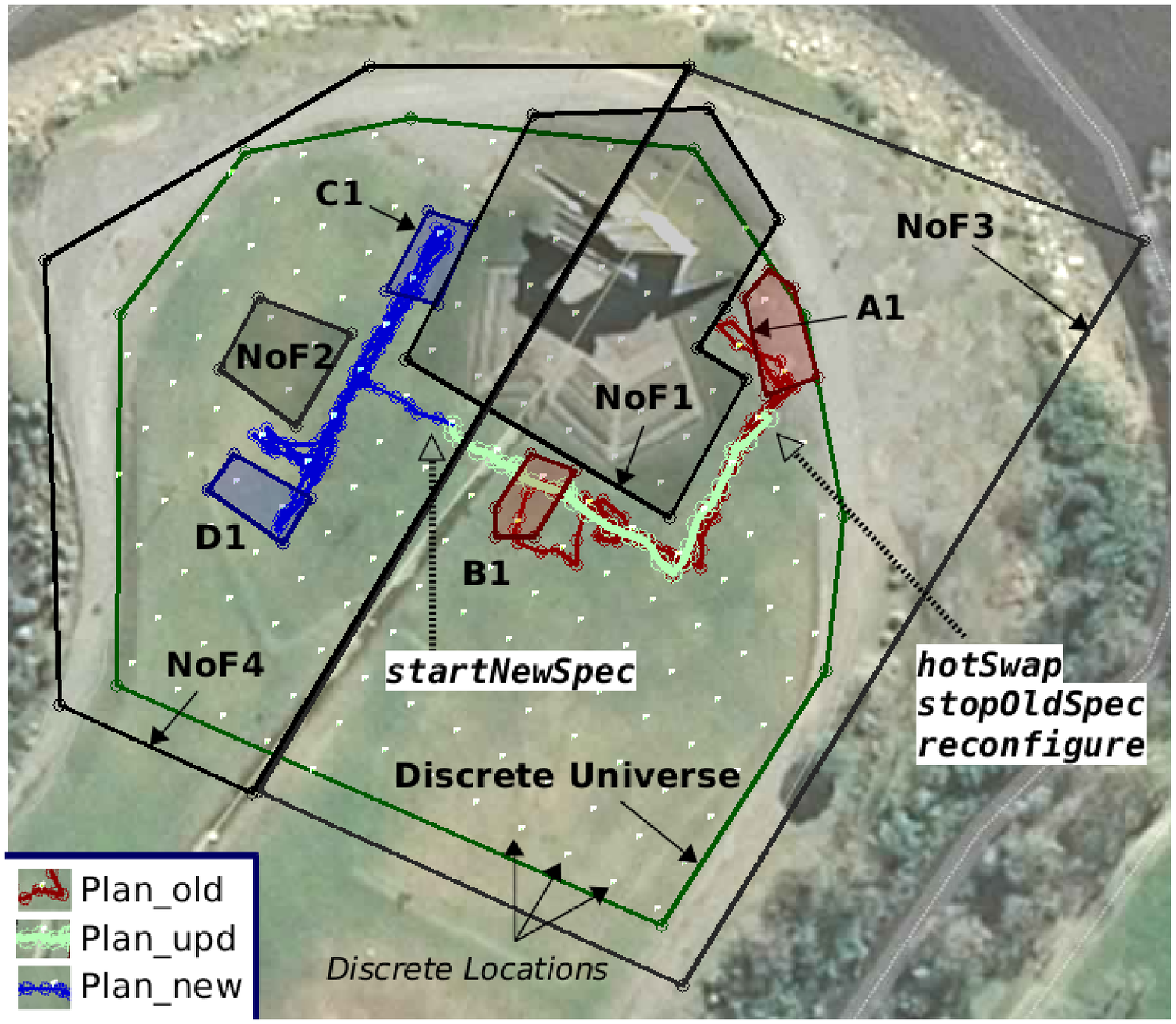}
 \caption{}
 \label{fig:patrol-patrol}
\end{subfigure}
~
\begin{subfigure} {0.335\linewidth}
 \centering
\includegraphics[width=1.0\linewidth]{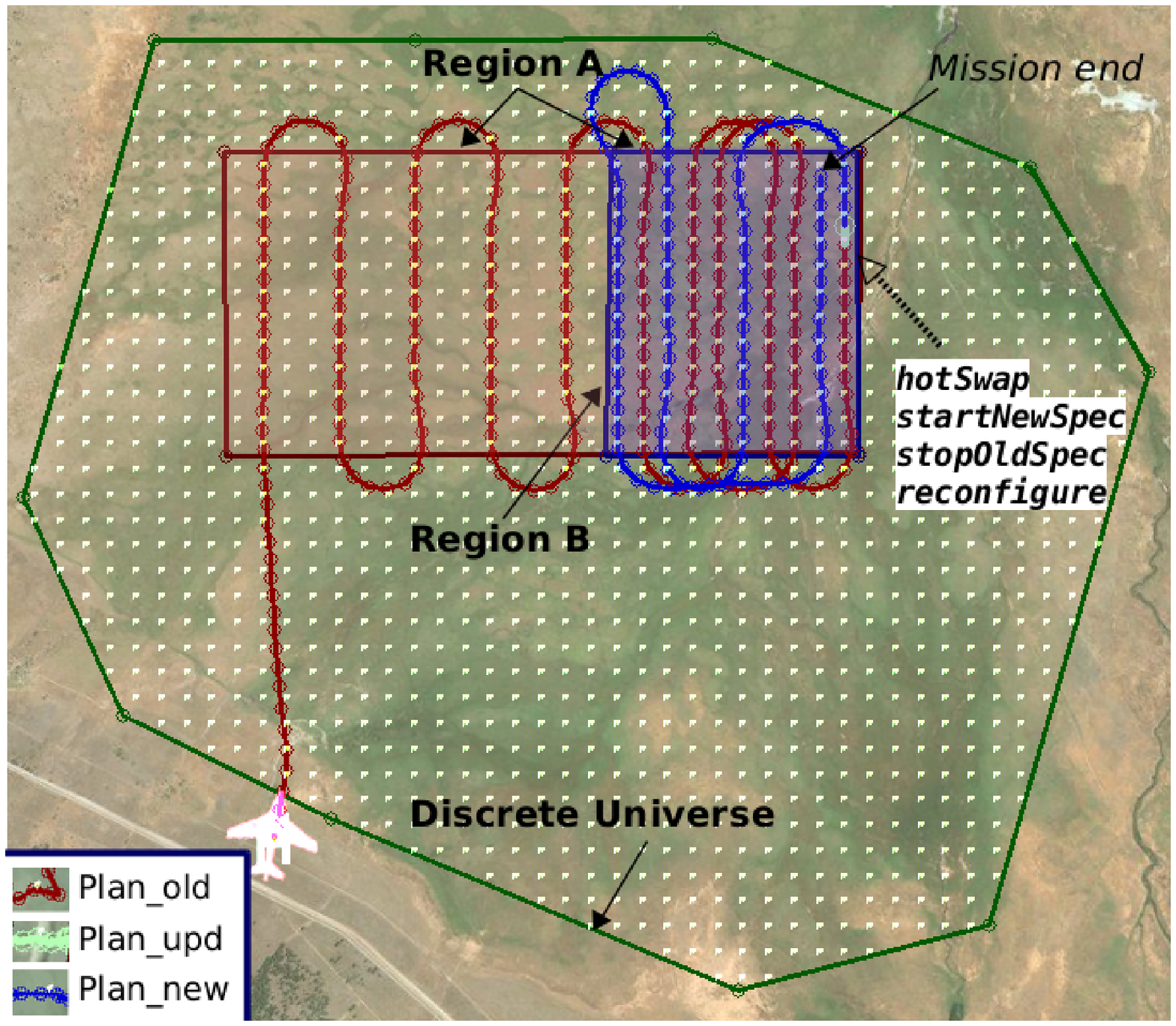}
 \caption{}
 \label{fig:cover-cover}
\end{subfigure}
~
\begin{subfigure} {0.32\linewidth}
\centering
\includegraphics[width=1.0\linewidth]{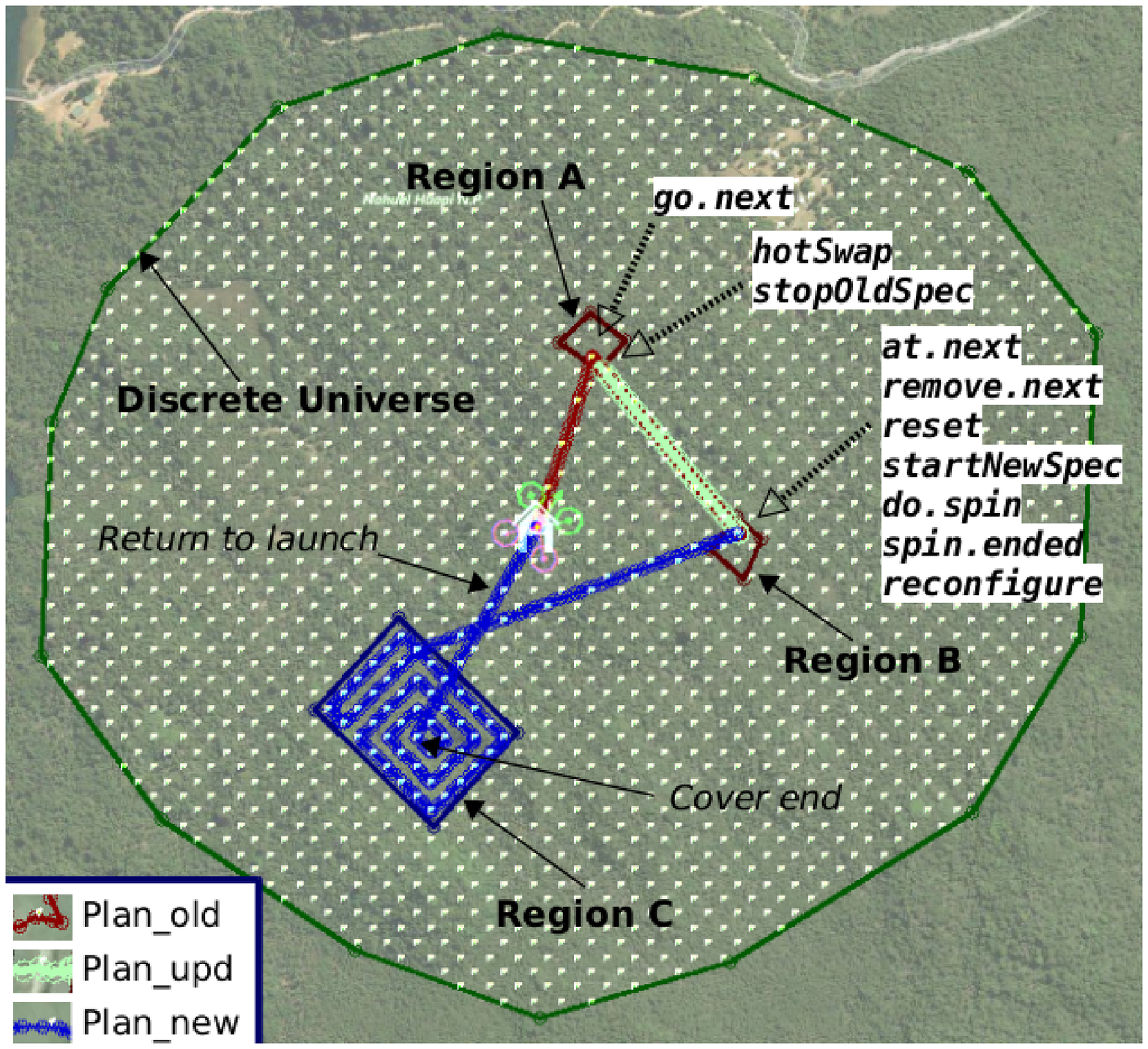}
 \caption{}
 \label{fig:firelookout}
\end{subfigure}
\caption{(a) Parrot Ar.Drone 2.0 real flight path with an unexpected goal change. (b) 
ArduPlane simulation for a mission degradation scenario. (c) ArduCopter simulation for a 
fire lookout and cover mission.}
\end{figure*}

In this section we report on various flights and adaptations we ran to validate our 
approach. We aimed to validate various characteristics of our adaptive system. 
Namely we pursued:

\begin{itemize}
 \item \emph{Feasability} by running multiple missions and informally validating that 
 the resulting UAV behaviour is consistent with the intended behaviour.  
This includes analysing if synthesis times and memory consumption are a problem 
for realistic missions.  
 
 \item \emph{UAV flexibility} by using different UAVs. We flew a Parrot Ar.Drone in 
Section~\ref{sec:unexgoalchange} and~\ref{sec:rescue}, and ArduPilot 
simulations for fixed-wing UAVs (ArduPlane) and quadcopters (ArduCopter) in 
Sections~\ref{sec:battery} and~\ref{sec:fire}, respectively.
 \item \emph{Hybrid Control Layer flexibility} by implementing two different 
abstraction approaches: Iterator-Based Planning (Sections~\ref{sec:battery} 
and~\ref{sec:fire}) and Explicit-Location Planning (Sections~\ref{sec:unexgoalchange} and 
\ref{sec:rescue}).
 \item  \emph{Mission variability} by studying different missions types with varying 
discrete universe sizes, ranging from \SI{48} to \SI{1834} discrete locations. Missions 
 patterns we used are common 
in the literature (see~\cite{Menghi19} for a survey). Within this item, we also looked 
at the ability of the system to support non-trivial \emph{reconfiguration} by 
introducing various types of new sensors. 
\end{itemize}

The videos and specifications for the simulated and real flights can be found 
in supplementary material.

\subsection{Unexpected Goal Change} \label{sec:unexgoalchange}

We  revisit the Example 3 of Section~\ref{subsec:trans}. The 
original mission consists of a typical patrol mission as in~\cite{Menghi19, Garcia2019, 
Yan17, Livingston13, Castro15} for surveillance of two areas \region{A1} and \region{B1} 
with two no-fly 
zones: \region{NoF1} to avoid a local obstacle in the fly region and \region{NoF4} as the 
\region{\nfold}. These areas are shown in Figure~\ref{fig:patrol-patrol}. For this mission 
we used discrete cells of $\SI{10}{\metre} \times \SI{10}{\metre}$ and a flight height of 
\SI{1.5}{\metre}, with a universe of \SI{163} discrete locations.

A discrete event controller can be synthesised using a similar explicit-location 
abstraction as in Figure~\ref{fig:LTS}, expanding the model to include takeoff 
and landing events, fluents defined as in Section~\ref{subsec:adaplive}, a safety rule 
$\Gl \neg \land$ to avoid unnecessary landing and the 
following requirement: $(\Gl \neg \arrivedfl{NoF1}) \wedge (\Gl \neg \arrivedfl{NoF4}) 
\wedge (\Gl \F \arrivedfl{A1}) \wedge (\Gl\F \arrivedfl{B1} )$.

A controller was synthesised in \SI{0.5}{\second} using up to \SI{20.1}{\mega \byte} 
of RAM 
and automatically loaded onto the Parrot Ar.Drone 2.0 
which started the mission and produced the trajectory indicated as \plan{old} in 
Figure~\ref{fig:patrol-patrol}. 
While flying the  \plan{old}, we specified a new goal: two new areas to be patrolled 
\region{C1} and \region{D1}, 
together with the no-fly regions \region{NoF2} due to local obstacles and \region{NoF3} as 
the \region{\nfnew}. To avoid inconsistency between the two missions, we added a 
transition requirement similar to (\ref{eqn:T-arrived45}) but prohibiting the local 
obstacles \region{NoF1} and \region {NoF2} instead of forcing the UAV to be in locations 
\region{4} 
or \region{5}.

The update controller that was synthesised in \SI{7.4}{\second} (using a maximum of 
\SI{46.7}{\mega \byte}) 
and 
uploaded, 
stops the old 
specification (\stopOld) while in region \region{NoF3}, but the new specification is 
started (\startNew) much later, only when the UAV leaves \region{NoF3} (see 
trajectory \plan{upd}), as it is prohibited in the new specification. The UAV then 
carries on its new patrol mission of regions \region{C1} and \region{D1} as seen in 
trajectory \plan{new}.

\subsection{Unexpected Battery Consumption Rate} \label{sec:battery}
We explore a different scenario of mission plan update: \emph{mission degradation 
due to unforseen circumstances}. 
We assume an original mission that requires covering an area 
\region{A} (i.e., visit every discrete location in \region{A}) for mapping purposes 
(e.g.,~\cite{Rokhmana16,Mapping,Cabreira19}).

In large discrete regions,  cover missions with explicit-location modelling do 
not scale well as one fluent for every location is needed to track if it has been 
covered. 
Instead, we use an alternative discrete abstraction strategy: Iterator-Based 
Planning~\cite{Iterator}. This  follows the idea of sensor-based planning 
(e.g.,~\cite{Gazit09}) in which a sensor (i.e., code) is provided to identify if a 
location correspond to the area to be covered. 

\begin{figure}
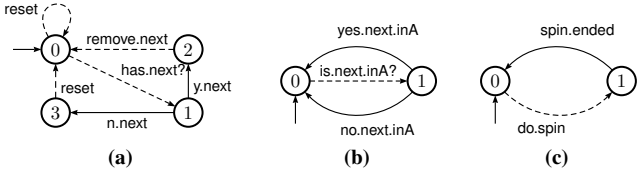

\begin{subfigure} {0.38\linewidth}
\centering
\input{src/lts/iterator}
\caption{}
\label{fig:iteratorLTS}
\end{subfigure}
~
\begin{subfigure} {0.27\linewidth}
\centering
\input{src/lts/SensorNextLTS}
\caption{}
\label{fig:sensorLTS}
\end{subfigure}
~
\begin{subfigure} {0.27\linewidth}
\centering
\input{src/lts/spinLTS}
\caption{}
\label{fig:spinLTS}
\end{subfigure}
~
\caption{(a) Iterator LTS. (b) Sensor A LTS. (c) Spin capability LTS.}
\end{figure}

Iterator-Based Planning works by providing a high-level iterator 
API (see Figure~\ref{fig:iteratorLTS}) that allows us to iterate over the discrete 
locations, abstracting from the number of locations involved. The user can then specify 
in 
LTL what must be done for each location as it is iterated over. For example, for a cover 
\region{A} mission, every time the iterator responds that there  is still a location to 
process (\yesNext), the controller should ask \inP, and it it is (\yinP) it should \go to 
that location (see Figure~\ref{fig:sensorLTS}). 

The system synthesised a controller for this mission in \SI{0.5}{\second} 
 (using up to \SI{14.0}{\mega \byte} RAM) and flew 
the UAV (we ran two 
missions, one with the Parrot 
Ar.Drone and the other  with a simulated ArduPlane SITL). For the plane, a 
region \region{A} was defined by the user as shown in Figure~\ref{fig:cover-cover}, 
discrete cells were of $\SI{60}{\metre} \times \SI{60}{\metre}$ and the flight height of 
\SI{100}{\metre}, generating a universe of \SI{1251} discrete locations. From region 
\region{A}, a sensor was automatically 
computed by the \swarch{Discretizer} and uploaded onto the UAV through the \swarch{Hybrid 
Module TX} before starting the original mission plan. The UAV produced
the trajectory \plan{old} shown in Figure~\ref{fig:cover-cover} while covering \region{A}.

Suppose that due to wind conditions or a malfunctioning engine, the battery consumption 
rate is higher than predicted by a runtime monitor as in~\cite{Wei18}. At the 
\swarch{Control Panel} level this could fire off an 
alarm indicating that the UAV will be unable to completely 
cover region \region{A}. We simulated this scenario and had the user intervene (with the 
UAV still in-flight) by producing a degraded mission plan: the user chooses to reduce to 
half the original region \region{A}, to at least have one contiguous region completely 
covered.  This was done first by defining~\region{B} in Figure~\ref{fig:cover-cover} and 
then specifying the new mission requirements, together with the standard transition 
requirement $\Theta_\emptyset$ and the rule~(\ref{eqn:T-initialA}) that only 
allows \reconfigure to happen when \sensor{A} is in its initial state. The 
environment state map $E\lig{g}{\reconfigure} E'$ was defined by ignoring the state of 
\sensor{A} and 
mapping all states from $E$ to the equivalent in $E'$ with \sensor{B} in its initial 
state.

The  \swarch{Discretizer} module automatically generates the Python code that 
implements the new \sensor{B} and uploads it onto the flying UAV. Meanwhile the 
synthesizer produced an update controller in \SI{1.8}{\second} (using up to 
\SI{16.2}{\mega \byte})
that, when uploaded and 
hot-swapped, stopped (\stopOld) the old mission, executed 
\reconfigure 
which triggers the binding of \sensor{B} module to the \swarch{Hybrid Control Layer} (and 
the 
unbinding of \sensor{A}), and signals the start of the new mission (\startNew).
\begin{equation}\label{eqn:T-initialA}
\begin{aligned}
 \Theta &= \Gl (\reconfigure \Rightarrow \neg \formatFluent{SensingA}) \\
 \formatFluent{SensingA} &= \langle \inP, \set{\yinP,\ninP} ,\bot \rangle
\end{aligned}
 \end{equation}
The UAV continues  its mission (Figure~\ref{fig:cover-cover}) covering \region{B} 
(\plan{new}).

\subsection{Fire Monitoring} \label{sec:fire}
UAVs are used  to aid firefighters by fire monitoring and 
tracking~\cite{FireMonitoring,Beachly17}. A fire monitoring mission can be as simple 
as a fire lookout~\cite{FireLookout} between two locations far apart from each other.
Such a mission is suitable for a multi-rotor with stationary flight capabilities.
The mission we simulated consists of visiting two areas \region{A} 
and 
\region{B}, doing a full slow spin (see LTS in Figure~\ref{fig:spinLTS}) at each of them 
to have a \SI{360}{\degree} view of the surrounding area. If the quadcopter has a camera 
mounted aboard and streams its video to a remote monitoring station, a human  
can view this footage to detect the presence of fire. We synthesised (in 
\SI{0.3}{\second} using \SI{15.8}{\mega \byte} of RAM) and ran this mission using a  
simulated ArduCopter SITL and an iterator-based planning approach with discrete 
cells of $\SI{30}{\metre} \times 
\SI{30}{\metre}$ and a flight height of \SI{70}{\metre}, totalling \SI{1834} discrete 
locations. The resulting trajectory can be seen as \plan{old} in 
Figure~\ref{fig:firelookout}.

Our adaptation is useful in this scenario if the human 
in-the-loop needs a closer look at a certain area where fire is suspected to 
be present. The user can then select a new region to cover and generate, with help of the 
\swarch{Discretizer}, the required \sensor{C} code. To point the camera aboard 
the UAV downward, the user modifies manually the generated code to move the 
camera controlling servo when the \swarch{Hybrid Module} is initialised. The new 
mission consists of covering the area \region{C} and to return to launch when 
finished.

The environment mapping is similar to the one in Section~\ref{sec:battery}: 
from $E$ (with \sensor{A} and \sensor{B}) to $E'$ (with \sensor{C}). The 
transition requirement is a combination of 
(\ref{eqn:T-initialA}) adapted to include \sensor{B} and the spin capability in their 
initial states, $\Theta_1$ and $\Theta_2$ to restrict actions between specifications, 
and $\Theta_3$ to force a reset of the iterator before \startNew
(to guarantee full coverage of the region \region{C}). 
\begin{equation}\label{eqn:T-fire}
\begin{aligned}
 &\Theta_1 = \Gl \big((\stopOldfl \wedge \neg \startNewfl) \Rightarrow \neg 
\senseormove \big) \\
 &\Theta_2 = \Gl (\reconfigure \Rightarrow \startNewfl) \\
 &\Theta_3 = \Gl (\startNew \Rightarrow \iterempty), \iterempty = \langle \reset, 
\hasNext, \bot \rangle
\end{aligned}
\end{equation}
where \senseormove is set to true with moving and 
sensing actions (e.g., \takeOff, \go, \inP) and false with the rest.

The update controller was synthesised in \SI{36.8}{\second} (\SI{35.1}{\mega \byte}) and 
uploaded 
immediately.
Figure~\ref{fig:firelookout} exemplifies well the non-trivial update strategy that the 
update controller had to execute to satisfy all requirements: The \hotSwap occurred 
while flying to a new location (i.e., just after \goN{next} but before \arrivedN{next}). 
The controller immediately does \stopOld but cannot do \startNew as $\iterempty$ 
does not hold (see $\Theta_3$). Further, to do \reset it must not be moving (see 
Iterator 
requirements~\cite{Iterator}), thus it first  waits until 
the new location is reached (which it can assume will eventually happen), then resets 
the 
iterator and then does \startNew and \reconfigure. Between \startNew and 
\reconfigure, 
the update controller chooses to do a spin (\spin) since it does not violate any 
requirement.


\subsection{Unexpected Search \& Rescue} \label{sec:rescue}
Search \& rescue scenarios are a common theme in robotics 
(e.g.,~\cite{Yamazaki19,Wang18,Gazit07}) and the flying of the example in 
Section~\ref{sec:intro} showcases the ability of our system to  adapt from a 
patrol to a search \& rescue mission including uploading of non-trivial functionality. 


The original mission is high-height patrol similar to the one in 
Section~\ref{sec:unexgoalchange}, synthesised in \SI{0.4}{\second} (using up to 
\SI{16.5}{\mega 
\byte}) for \SI{48} discrete locations. We flew the synthesised controller using 
the Parrot Ar.Drone 2.0. The mission is then updated into a low-height flight of the 
same patrol locations but introducing an 
image processing \swarch{Hybrid 
Module} to sense at each arrived 
location for red objects and the 
 following specification, where \photofl is a fluent that is true 
when the 
image processing module detects a red object:
$$\forall 0 \leq i, j \leq 47 \cdot \big[i \neq j 
\wedge 
\Gl \big(\arrivedfl{i} \rightarrow (\neg \arrivedfl{j} \W \photofl) \big)\big]$$

This image processing code is 
uploaded onto the UAV prior to the \hotSwap command being issued.
The height inconsistency (high- vs low-height) is solved by the following 
transition specification: $\Gl \big( (\stopOldfl \wedge \neg \startNewfl) \implies \neg 
(\Gamma \setminus \{\lowHeight,$ $\highHeight \}) 
\big)$, where the set $\Gamma$ holds all the controllable 
actions. Update synthesis time totalled \SI{15}{\second} (\SI{54.6}{\mega \byte}).

\section{Discussion and Future Work}
\label{sec:discussion}

We believe that the experimental results provide some evidence that runtime synthesis can 
be used to support mission adaptation in real UAV systems. Of course, there are 
threats to validity. The main one being that experimental results may not generalise 
to all UAV setups and vehicle configurations. Certainly, one limitation is that 
synthesis is being applied to a single UAV and that we are considering only atomic 
reconfiguration to simplify implementation and presentation (i.e., reconfiguration 
strategies as in~\cite{Tajalli2010} are not supported). 

Many missions of interest in this domain 
are multi-vehicle~\cite{Cabreira19}. One potential limitation of the approach is that 
discrete event controller synthesis may not scale to large missions. However, 
techniques such as Iterator-Based  Planning~\cite{Iterator} have shown that 
missions for hundreds of thousands of discrete locations are tractable. Although we 
extended DCU for the purpose of typical recurring requirements in 
robotic missions, a comprehensive study of extensions to DCU to account for more 
complex liveness requirements is still needed and would be interesting to develop in 
future work.
\section{Conclusions}

We have presented a novel architecture for UAV systems that supports correct by 
construction mission adaptation performing synthesis of discrete event controllers 
at runtime and hot-swapping them onto a UAV. The architecture 
supports both behavioural and structural adaptation building on hybrid 
control, dynamic controller update and adaptive 
software architectures.
We show in several 
missions taken from the robotic literature that new mission goals can be 
introduced and correctly updated into a running system, both for real and simulated 
scenarios. Having shown how the update 
problem is non-trivial, we demonstrate how user specified transition requirements 
can be used to solve inconsistencies and correctly synthesise update strategies, that 
are guaranteed to take the  running system into a state where the software 
architecture can be reconfigured and the new plan can be executed.


\bibliographystyle{IEEEtran}
\bibliography{IEEEabrv,src/biblio}

\end{document}